%% file: neurips_2026.tex
\documentclass{article}

 \usepackage[preprint]{neurips_2026}

\usepackage[utf8]{inputenc} %
\usepackage[T1]{fontenc}    %
\usepackage{hyperref}       %
\usepackage{url}            %
\usepackage{booktabs}       %
\usepackage{graphicx}       %
\usepackage{amsmath}        %
\usepackage{amsfonts}       %
\usepackage{amssymb}        %
\usepackage{nicefrac}       %
\usepackage{microtype}      %
\usepackage{xcolor}         %
\usepackage{wrapfig}
\usepackage{colortbl}
\PassOptionsToPackage{numbers}{natbib}
\title{SpatialAvatar-0: High-Quality 4D Head Avatar with Multi-Stage Reconstruction}

\author{
Yiran Wang$^{1*}$\quad
Zeyu Zhang$^{2*}$\quad
Yuanming Li$^{2*}$\quad
Ziming Wang$^{3}$\quad
Yang Zhao$^{4\dag}$
\\
[0.3em]
$^1$USYD\quad
$^2$SpatialReal\quad
$^3$ZJU\quad
$^4$La Trobe\\
[0.1em]
\footnotesize $^*$Equal contribution.
$^\dag$Corresponding author: y.zhao2@latrobe.edu.au.
}

\begin{document}

\maketitle

\begin{abstract}
High-quality 4D head avatars from one or a few source portraits are central to telepresence, AR/VR, and digital-human interaction. 3D Gaussian Splatting (3DGS) has emerged as the dominant representation, with two complementary regimes (generalizable feed-forward predictors and per-subject refiners) maturing in parallel.
However, existing feed-forward predictors are trained on a single dataset family with a hard-coded source count, inheriting the corresponding domain bias. Per-subject refiners require $300$K--$600$K iterations and rely on adaptive densification that destroys upstream Gaussian layouts, preventing the two regimes from sharing a representation end-to-end.
To bridge both regimes we propose \textbf{SpatialAvatar-0} on a shared FLAME-mesh-bound Gaussian representation: a feed-forward generator with a parameter-free $K$-source mean-pool and a monocular-temporal $\to$ multi-view-spatial two-phase schedule that anchors against identity-prior collapse onto the smaller multi-view set. We further introduce a $10$K-iter layout-preserving per-subject refinement loop that freezes the FLAME-binding and Gaussian count and replaces densification with a three-component anti-spike regularization. On VFHQ/HDTF cross-domain zero-shot we surpass the in-domain leader GAGAvatar by $\mathbf{{+}1.5}$\,dB PSNR despite never training on either test domain, and on the SplattingAvatar monocular benchmark we lead every reported metric, surpassing the $300$K-iter GeoAvatar by $\mathbf{{+}1.3}$\,dB PSNR at up to $\mathbf{60{\times}}$ shorter per-subject schedule than common SOTA baselines.
Website:~\url{https://spatialwalk.github.io/SpatialAvatar-0}
\end{abstract}

\section{Introduction}

High-quality 4D head avatars are a building block for telepresence,
AR/VR communication, and digital-human interaction. Within the explicit
3D Gaussian Splatting (3DGS) line of head-avatar
research~\citep{kerbl20233d}, two complementary regimes have matured:
\emph{generalizable feed-forward predictors}~\citep{chu2024generalizable,
chu2024gpavatar,deng2024portrait4dv2,ye2024real3dportrait} that emit an
animatable face-bound Gaussian model in a single forward pass from one
or a few source portraits, and \emph{per-subject 3DGS
refiners}~\citep{qian2024gaussianavatars,xiang2024flashavatar,
moon2025geoavatar,shao2024splattingavatar} that continue to optimize a
face-bound Gaussian model against a held-out video of a single identity
to recover subject-specific high-frequency detail. Production
deployments of digital humans need both regimes, so the practical
question has shifted from ``feed-forward or per-subject'' to whether
the two regimes can share a common Gaussian representation end-to-end.

Two long-standing gaps prevent such a unified pipeline. \emph{First},
existing feed-forward Gaussian avatars are trained on a single dataset
family (either monocular video~\citep{zhu2022celebvhq} or multi-view
capture~\citep{kirschstein2023nersemble}) and inherit the
corresponding domain bias: monocular-trained predictors lack
ground-truth multi-view geometry, while multi-view-trained predictors
see orders-of-magnitude fewer identities and degrade on in-the-wild
portraits. Architectures further hard-code the source-image count,
preventing a single network from leveraging however many source views
happen to be available at deployment time. \emph{Second}, per-subject
3DGS refinement consumes $300\text{K}$--$600\text{K}$ iterations per
identity (${\sim}5$--$9$ hours)~\citep{xiang2024flashavatar,
qian2024gaussianavatars,moon2025geoavatar} and relies on adaptive
densification that destroys any spatial inductive bias inherited from a
feed-forward initialization. Removing densification, however, exposes
the well-known 3DGS anisotropy failure: Gaussians elongate along a
single axis to overfit individual training views, catastrophically
breaking novel-view rendering.

We observe that a Gaussian avatar trained on the \emph{union} of
monocular and multi-view supervision can in principle inherit
wide-identity coverage from monocular pretraining and geometric
grounding from multi-view post-training, provided the training signal
does not collapse the broader identity prior onto the smaller
multi-view set; we further want the same network to handle a variable
source count without architectural switching. Separately, we want a
per-subject refinement loop that is \emph{layout-preserving}: it
freezes the FLAME-binding and Gaussian count of the feed-forward
output while refining per-Gaussian attributes, so that the upstream
identity prior remains reusable downstream and the refinement consumes
orders of magnitude fewer iterations, provided we can suppress the
3DGS anisotropy spikes that emerge once densification is removed.

We propose \textbf{SpatialAvatar-0}, a unified feed-forward and
per-subject 4D head-avatar pipeline over a shared FLAME-mesh-bound
Gaussian representation. \emph{First}, addressing the feed-forward
gap, we introduce a feed-forward image-to-Gaussian generator with a
parameter-free $K$-source mean-pool aggregator that is the identity at
$K{=}1$ and order-invariant for $K{>}1$, trained under a
monocular-temporal $\to$ multi-view-spatial two-phase schedule with an
L2-SP anchor~\citep{li2018l2sp} and a $25\%$ NeRSemble cross-time mix
that together prevent identity-prior collapse onto the smaller
multi-view set. \emph{Second}, addressing the per-subject gap, we
propose a $10$K-iter layout-preserving per-subject refinement loop
that freezes the FLAME-triangle binding and the Gaussian count and
replaces densification with a three-component anti-spike regularization
(scale-freeze warmup, hard log-scale clamp, screen-space
anti-anisotropy penalty) that keeps the maximum Gaussian aspect ratio
bounded throughout optimization. \emph{Third}, we conduct a
comprehensive empirical evaluation: on cross-domain VFHQ and HDTF
feed-forward zero-shot probes our model surpasses the in-domain leader
GAGAvatar~\citep{chu2024generalizable} by $\mathbf{{+}1.5}$\,dB PSNR
despite never observing either dataset during training; on the
SplattingAvatar monocular benchmark our $10$K-iter
refinement surpasses the $300$K-iter
GeoAvatar~\citep{moon2025geoavatar} on every reported metric by
$\mathbf{{+}1.3}$\,dB PSNR while completing per-subject creation in
${\sim}2$\,minutes (vs.\ $4.9$\,hours); ablations confirm each design
choice is load-bearing.

Our contributions are threefold:
\begin{itemize}
\item A $K$-source-variable feed-forward FLAME-mesh-bound Gaussian generator
with a monocular $\to$ multi-view two-phase training schedule, anchored
by L2-SP and a $25\%$ NeRSemble cross-time mix against identity-prior
collapse on the smaller multi-view set.
\item A $10$K-iter layout-preserving per-subject refinement loop with a
three-component anti-spike regularization replacing densification, leading
every reported metric on the SplattingAvatar leaderboard at up to
$\mathbf{60{\times}}$ shorter schedule than common SOTA per-subject
baselines.
\item Comprehensive cross-domain and per-subject experiments on VFHQ,
HDTF, and the SplattingAvatar monocular benchmark
(\S\ref{sec:experiments}), with ablations validating every design
choice.
\end{itemize}

\section{Related Work}

The creation of photorealistic and animatable head avatars has {garnered significant attention} within the computer vision and graphics communities, driven by applications in telepresence, gaming, and the metaverse. The fundamental objective of this field is to faithfully reconstruct or synthesize a source head from sparse inputs, such as a single portrait or a short monocular video, while enabling precise, fine-grained control over facial expressions, gaze, and head poses. Existing methodologies can be broadly categorized into 2D-based synthesis, implicit 3D reconstruction using radiance fields, and the emerging paradigm of explicit 3D Gaussian Splatting.

\subsection{2D-Based Talking Head Synthesis}

Early attempts to generate talking heads primarily relied on 2D generative models, such as GANs, to synthesize image sequences directly in the pixel space. A popular strategy among these works is the injection of expression and pose features from a driving image into a 2D generative backbone to achieve motion through feature modulation or latent space manipulation~\cite{zakharov2019few,burkov2020neural,  wang2023progressive, ma2023otavatar, ma2023dreamtalk, ji2021audio, kou2020talking}. Another recent trend involves the estimation of dense 2D warp fields or optical flows to deform the source portrait into the target geometry~\cite{siarohin2019first, ren2021pirenderer, drobyshev2022megaportraits, hong2022depth, zhang2023metaportrait, siarohin2023unsupervised, zhang2021flow}. To improve the physical realism and structural integrity of these deformations, several methods have integrated 3D Morphable Models (3DMMs)~\cite{blanz1999morphable, paysan20093d, li2017learning} to serve as a low-dimensional geometric prior~\cite{wang2021oneshot, yin2022styleheat, drobyshev2022megaportraits, zhang2023metaportrait, ma2023otavatar, sanyal2019learning}. Some works further leverage the generative power of StyleGAN to produce high-resolution textures conditioned on 3DMM coefficients~\cite{sun2023next3d, yin2023nerfinvertor}. {Although these methods produce visually compelling results}, they {inherently struggle with} 3D consistency, often leading to unrealistic distortions, temporal flickering, and identity-shifting artifacts when encountering significant head pose variations. Furthermore, these approaches often {lack explicit 3D geometry constraints}, limiting their utility in free-viewpoint rendering and complex lighting interaction applications.

\subsection{3D-Based Head Avatar Reconstruction}

To overcome the view-consistency limitations of 2D methods, a {paradigm shift} toward 3D-aware representations has occurred. Early 3D approaches utilized mesh-based modeling driven by statistical face priors to provide explicit surfaces for rendering~\cite{khakhulin2022realistic, grassal2022neural, wang2024prior, feng2021learning, lattas2023fitme}. However, the rise of Neural Radiance Fields (NeRF)~\cite{mildenhall2020nerf} has enabled more flexible, topology-agnostic head reconstruction. Numerous NeRF-based methods have been proposed to reconstruct personalized avatars from monocular or sparse-view videos by learning a canonical radiance field combined with a deformation field~\cite{gafni2021dynamic, athar2022rignerf, gao2022personalized, park2021nerfies, zielonka2023insta, kirschstein2023nersemble, bai2023personalized, zhao2023havatar, athar2023flame}. To facilitate one-shot synthesis and improve cross-identity robustness, researchers have explored learning tri-plane features or conditioned latent spaces to bypass the need for subject-specific training~\cite{li2023generalizable, ma2023otavatar, chu2024gpavatar, ye2024geneface, deng2024portrait4d, taubner2025cap4d, li2023hide}. {Despite their significant progress}, NeRF-based avatars typically suffer from {face rendering speed limitations} due to the heavy computational cost of volumetric sampling through large MLPs, making them difficult to deploy in real-time environments. Moreover, many of these methods {depend on identity-specific optimization} or require thousands of frames for training, which significantly limits their generalization to unseen identities and poses significant privacy and storage concerns.

\subsection{3D Gaussian Splatting for Human Avatars}

Recently, 3D Gaussian Splatting (3DGS)~\cite{kerbl20233d} has emerged as a revolutionary representation, offering an optimal {balance between rendering efficiency and geometric fidelity}. Unlike implicit NeRFs, 3DGS utilizes anisotropic explicit primitives that allow for real-time rasterization via a tile-based approach. Recent research has successfully extended 3DGS to human head modeling by binding Gaussians to underlying meshes or learned deformation fields~\cite{shao2024splattingavatar, qian2024gaussianavatars, lee2024surfhead, zhao2024psavatar, hu2024gaussianavatar, zhang2025hravatar, xu2023gaussian}. Some approaches map Gaussians onto UV coordinates~\cite{xiang2024flashavatar, jiang2024uv} or utilize neural parametric models to handle dynamic expressions and complex topologies~\cite{giebenhain2024npga, xu2023gaussian}. Most recently, generalizable one-shot Gaussian avatars have been proposed to enable immediate animation from a single image by predicting Gaussian attributes directly from pixel features~\cite{chu2024generalizable, zheng2024headgap, lyu2024facelift, gao2026sega, moon2025geoavatar, cai2025hybrid}. {However}, current Gaussian-based methods often {lack robust generalization} across diverse accessories (e.g., glasses, hats) and complex hairstyles, as they are frequently trained on datasets with limited diversity. Furthermore, many existing Gaussian avatars still {require identity-specific video data} to achieve high-frequency textural details, failing to bridge the gap toward a truly generalizable, zero-shot digital human system that achieves ultra-realistic results across the entire human population.

In contrast to these prior works, our method leverages a hybrid representation that {bypasses the redundant computations} of volumetric rendering while maintaining superior generalization capabilities. By integrating a multi-scale feature fusion module with an enhanced rigging strategy, we achieve high-fidelity, view-consistent rendering without the need for time-consuming per-identity optimization.

\input{method}

\input{experiment}

\section{Conclusion}

We presented \textbf{SpatialAvatar-0}, a unified end-to-end pipeline
that bridges generalizable feed-forward and per-subject 3DGS
head-avatar regimes through a shared FLAME-mesh-bound Gaussian
representation. The $K$-source-variable feed-forward generator combined
with a monocular $\to$ multi-view two-phase schedule produces a strong
cross-domain identity prior that surpasses the in-domain leader by
${+}1.5$\,dB PSNR on zero-shot VFHQ/HDTF benchmarks; the $10$K-iter
layout-preserving per-subject refinement, enabled by an anti-spike
regularization replacing adaptive densification, surpasses the
$300$K-iter GeoAvatar leaderboard on every reported metric by
${+}1.3$\,dB PSNR at $60{\times}$ shorter per-subject wall-clock.
Together these designs
demonstrate that the two avatar regimes need not be developed in
isolation: a coherent shared Gaussian representation, paired with
appropriate cross-domain training and layout-preserving refinement, can
deliver feed-forward generalization and per-subject fidelity within a
single ${\sim}2$-minute pipeline.

\clearpage

\bibliographystyle{plain}
\bibliography{main}

\clearpage
\appendix

\input{method_appendix}

\end{document}

%% file: method.tex
\section{Method}
\label{sec:method}

\subsection{Overview}
\label{sec:method-overview}

\begin{figure}[htbp]
  \centering
  \includegraphics[width=0.95\textwidth]{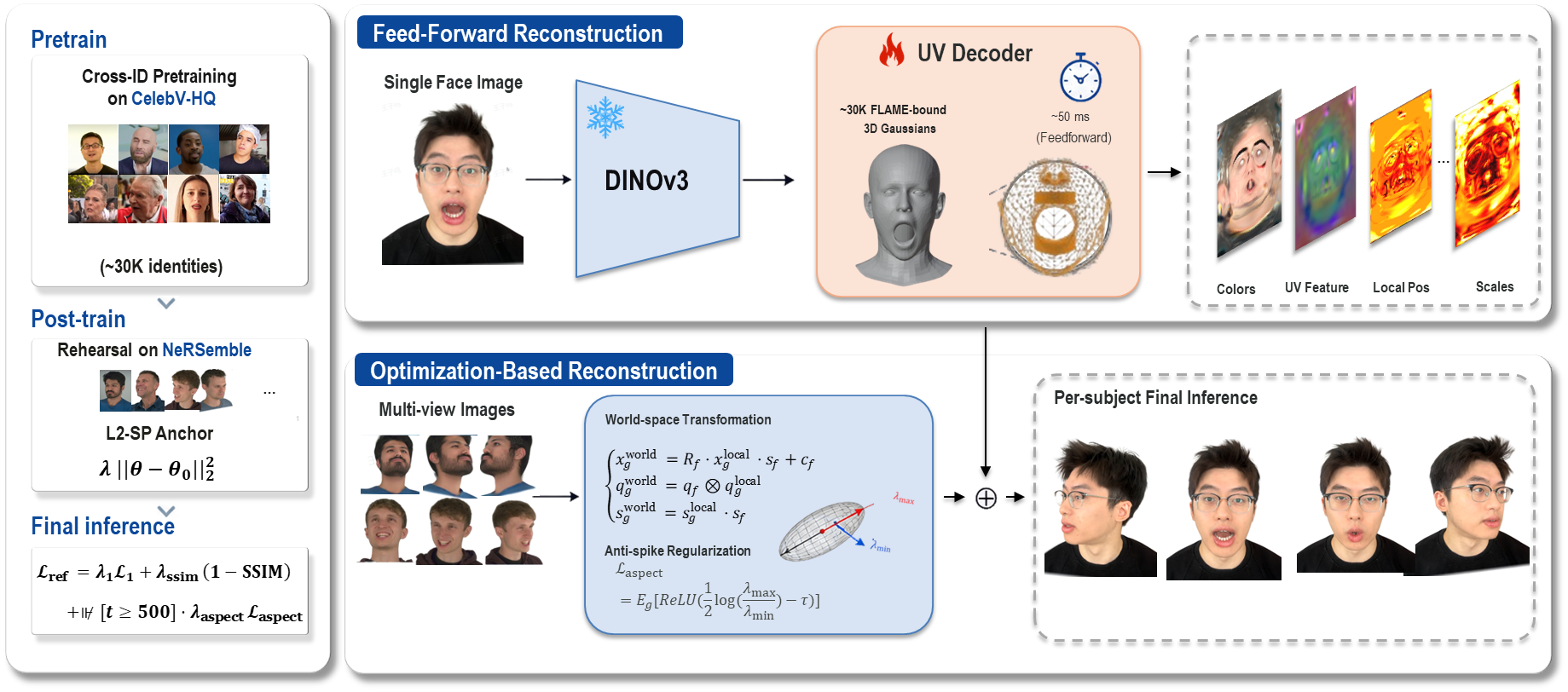}
  \caption{Overview of our two-stage avatar reconstruction pipeline.}
  \label{fig:method_overview}
  \vspace{-0.5cm}
\end{figure}

We frame head avatar reconstruction as two coupled stages (Fig.~\ref{fig:method_overview}) over a shared FLAME-mesh-bound 3D Gaussian representation~\citep{qian2024gaussianavatars}. \textbf{Stage 1} (feed-forward): a single network $f_\theta$ ingests $K\!\in\!\{1,2,3,4\}$ source portrait images and emits face-bound 3D Gaussians in one forward pass. \textbf{Stage 2} (optional per-subject optimization): starting from $f_\theta$'s output for a chosen reference frame, we run $10$K iterations of photometric refinement against the target video. Stage 1 is trained in two phases on the same architecture: monocular-temporal pretraining (Phase~1, CelebV-HQ~\citep{zhu2022celebvhq}) and multi-view-spatial post-training (Phase~2, NeRSemble~\citep{kirschstein2023nersemble}); the variable source count $K\!\in\!\{1,2,3,4\}$ during training exposes $f_\theta$ to monocular and multi-view contexts within a single training distribution.

\paragraph{Design summary.} We organize Gaussians via the FLAME UV unwrap: one Gaussian per valid pixel of a $256\!\times\!256$ UV grid ($\sim\!58$K per identity), each rigidly bound to its parent triangle. UV-aligned FLAME-bound Gaussian organizations have appeared in prior head-avatar work~\citep{xiang2024flashavatar}; we treat this layout as fixed infrastructure rather than as a contribution. Our paper-specific contributions are (i)~a feed-forward image-to-Gaussian generator over this layout with variable source count $K\sim\text{Uniform}\{1,2,3,4\}$, (ii)~a monocular-temporal $\to$ multi-view-spatial two-phase schedule with an L2-SP anchor against the Phase-1 checkpoint to prevent identity-prior collapse onto the smaller multi-view set, and (iii)~a per-subject refinement loop that freezes the binding manifold and the Gaussian count, preserving the one-to-one UV-pixel-to-Gaussian map of Stage~1. Comparison with feed-forward and per-subject head-avatar baselines is deferred to Related Work; quantitative comparisons appear in Experiments.

\subsection{Feed-Forward Reconstruction Stage}
\label{sec:method-feedforward}

\paragraph{Architecture.}
A frozen DINOv3 ViT-B/16 backbone with a trainable DPT head~\citep{ranftl2021vision} produces a dense feature map and a global CLS token; the dense feature is barycentrically warped into UV space and passed to a StyleUNet UV generator~\citep{wang2023styleavatar} (StyleGAN2-based~\citep{karras2020analyzing}, with Spatial-Feature-Transform layers modulated by the source CLS). The $32$-channel UV output is concatenated with a $27$-d harmonic encoding $\gamma(\mathbf{d})$ of the target camera direction $\mathbf{d}$ and decoded by five parallel heads (local position, rotation, scale, color, opacity). For $K\!>\!1$ sources we mean-pool the per-source UV features and CLS tokens (parameter-free, identity at $K{=}1$, order-invariant; attention-pool baseline in Tab.~\ref{tab:ablation}). FLAME parameters are recovered with the GAGAvatar tracker~\citep{chu2024generalizable}; full architecture, tracker stack, harmonic embedding form, and barycentric warp are in App.~\ref{sec:method-app-arch}.

\paragraph{FLAME-conditioned residual head.}
A FiLM~\citep{perez2018film}-modulated residual head adds per-attribute corrections to position, rotation, scale, and color (opacity excluded) using a $112$-d conditioning vector formed from the target frame's expression, pose, and eye codes. Shape $\boldsymbol{\beta}$ is intentionally excluded: it feeds mesh geometry directly through FLAME's blendshape basis and is the load-bearing axis of cross-subject generalization. The FiLM and residual layers are zero-initialized so the residual contributes nothing at start. Conditioning-vector breakdown, the rationale for excluding shape and view-direction from the residual head, and ablation handles are in App.~\ref{sec:method-app-arch}.

\paragraph{Mesh binding.}
We adopt the rigid mesh-binding of \cite{qian2024gaussianavatars} (\S3.2). For each Gaussian we construct the local frame $(\mathbf{c}_f, \mathbf{R}_f, s_f)$ of its bound triangle on the target FLAME mesh (full construction in App.~\ref{sec:method-app-arch}); the world-space attributes of Gaussian $g$ bound to triangle $f(g)$ are
\begin{equation}
\mathbf{x}^{\text{world}}_g = \mathbf{R}_{f(g)}\,\mathbf{x}^{\text{local}}_g \cdot s_{f(g)} + \mathbf{c}_{f(g)}, \quad \mathbf{q}^{\text{world}}_g = \mathbf{q}_{f(g)} \otimes \mathbf{q}^{\text{local}}_g, \quad \boldsymbol{s}^{\text{world}}_g = \boldsymbol{s}^{\text{local}}_g \cdot s_{f(g)},
\label{eq:face-binding}
\end{equation}
where $\otimes$ denotes Hamilton-product quaternion multiplication. We modify the binding by indexing Gaussians via valid UV pixels rather than per-triangle (eliminating split/clone densification) and by freezing the binding during the per-subject refinement (\S\ref{sec:method-optimization}). Rendering deforms the mesh to the target FLAME pose, transforms each Gaussian via Eq.~\ref{eq:face-binding}, and rasterizes through a differentiable 3DGS splatter~\citep{kerbl20233d}.

\paragraph{Training objective.}
\label{sec:method-objective}
We supervise the rendered image $\hat{\mathbf{I}}$ against the target $\mathbf{I}$ with
\begin{equation}
\mathcal{L} = \underbrace{\lambda_1 \mathcal{L}_1 + \lambda_{\text{ms}} \mathcal{L}_{\text{ms}} + \lambda_{\text{lp}} \mathcal{L}_{\text{lp}}}_{\text{photometric}} + \underbrace{\lambda_{\text{box}} \mathcal{L}_{\text{box}}}_{\text{face-region}} + \underbrace{\lambda_{\text{ps}} \mathcal{L}_{\text{ps}} + \lambda_{\text{ss}} \mathcal{L}_{\text{ss}}}_{\text{surface smoothness}} + \underbrace{\lambda_{\text{jac}} \mathcal{L}_{\text{jac}} + \lambda_{\delta} \mathcal{L}_{\delta}}_{\text{architectural}}.
\label{eq:loss-ff}
\end{equation}
The photometric trio is pixel-wise $L_1$, $1-\text{MS-SSIM}$~\citep{wang2003multiscale}, and LPIPS~\citep{zhang2018unreasonable}; $\mathcal{L}_{\text{box}}$ is an additional $L_1$ on the face bounding box; $\mathcal{L}_{\text{ps}}$ and $\mathcal{L}_{\text{ss}}$ are UV-domain TV penalties on local position and log-scale; $\mathcal{L}_\delta$ is an output-space $L_1$ shrinkage on the residual head's deltas (complementing parameter-space weight decay); $\mathcal{L}_{\text{jac}}$ enforces view-direction invariance on geometric heads (next paragraph). Explicit forms, the data-difficulty source-frame sampler, and final $\lambda_*$ values are in App.~\ref{sec:method-app-training}.

\paragraph{View-direction invariance via Jacobian penalty.}
The harmonic view-direction input lets color and opacity heads model view-dependent appearance, but would also permit the decoder to leak view direction into geometric attributes. We enforce per-pixel view-direction invariance on the local-frame, pre-projection position and scale predictions via a Hutchinson-randomized Frobenius Jacobian penalty: with i.i.d.\ Rademacher $\mathbf{v}_p\!\in\!\{-1,+1\}^{d_{\text{out}}}$ and $g(\gamma)\!=\!\sum_p\!\langle\mathbf{v}_p, f_{\text{geo},p}(\gamma)\rangle$, one has $\mathbb{E}_{\mathbf{v}}[\|\nabla_\gamma g\|_2^2]\!=\!\sum_p\|J_p\|_F^2$ with $J_p\!=\!\partial f_{\text{geo},p}/\partial \gamma(\mathbf{d})$ (proof in App.~\ref{sec:method-app-jacobian}). The penalty is
\begin{equation}
\mathcal{L}_{\text{jac}} = \log\!\big(1 + \|\nabla_\gamma g_{\mathbf{x}}\|_2^2 + \|\nabla_\gamma g_{\boldsymbol{s}}\|_2^2\big),
\label{eq:loss-jac}
\end{equation}
costing one backward pass against $\gamma(\mathbf{d})$ per iteration. The estimator follows the single-backward Jacobian-regularization scheme of \cite{hoffman2019robust} adapted to the UV-pixel setting. The penalty does not eliminate rasterizer-induced perspective foreshortening, nor does it address pose-conditioning leakage through the residual head (head pose and source camera direction are coupled in monocular tracker outputs). Variance and bias analysis, head-selection rationale, and the rationale for the $\log(1+\cdot)$ wrapping are in App.~\ref{sec:method-app-jacobian}.

\paragraph{Two-phase training.}
\textbf{Phase~1 (monocular-temporal):} we pretrain $f_\theta$ on CelebV-HQ; from each clip we sample $K\!\sim\!\text{Uniform}\{1,2,3,4\}$ frames as sources and a held-out frame as target. Phase~1 covers a wide identity manifold but provides no ground-truth multi-view geometry. \textbf{Phase~2 (multi-view-spatial):} we continue from the Phase-1 checkpoint on NeRSemble with no architectural change; each batch samples $K\!+\!1$ cameras from the synchronized 16-camera capture, mixing cross-camera and cross-time draws. Phase~2 uses encoder layer-wise LR decay and an L2-SP anchor~\citep{li2018l2sp} against the Phase-1 weights to prevent identity-prior collapse onto the smaller multi-view set. Optimizer schedule, sampler details, the cross-camera/cross-time mixing ratio, and the L2-SP weight are in App.~\ref{sec:method-app-training}.

\subsection{Optimization-Based Reconstruction Stage}
\label{sec:method-optimization}

\paragraph{Per-subject refinement.}
The feed-forward stage cannot recover subject-specific high-frequency content beyond the FLAME mesh's parametric range. We therefore run a $10$K-iter per-subject photometric refinement starting from $f_\theta$'s output at a chosen reference frame, with the FLAME mesh and triangle bindings frozen (preserving the one-to-one UV-pixel-to-Gaussian map of Stage~1). At each iteration, we sample a random target frame, deform the FLAME mesh, transform every Gaussian via Eq.~\ref{eq:face-binding}, and render. We deliberately omit the adaptive densification of GaussianAvatars~\citep{qian2024gaussianavatars} so as to preserve the spatial inductive bias of our representation. Reference-frame selection (a jaw-pose-magnitude heuristic with a noise-threshold fallback) and per-parameter learning rates are in App.~\ref{sec:method-app-opt}.

\paragraph{Anti-spike regularization.}
Without densification, the optimization is susceptible to the well-known 3DGS anisotropy failure~\citep{kerbl20233d,qian2024gaussianavatars}: Gaussians elongate along a single axis to overfit individual training views, degrading reconstruction at non-training views. We combine three mechanisms (full forms in App.~\ref{sec:method-app-opt}): a scale-freeze warmup for the first $500$ iterations, a hard log-scale clamp applied per iteration after scale unfreezes, and a soft screen-space anti-anisotropy penalty active from iteration $500$:
\begin{equation}
\mathcal{L}_{\text{aspect}} = \mathbb{E}_g\!\left[\mathrm{ReLU}\!\left(\tfrac{1}{2}\log\!\frac{\lambda_{\max}\!\big(\Sigma^{\text{2D}}_g\big)}{\lambda_{\min}\!\big(\Sigma^{\text{2D}}_g\big)} - \tau\right)\right],
\label{eq:loss-aspect}
\end{equation}
where $\Sigma^{\text{2D}}_g$ is the projected 2D covariance of Gaussian $g$ at the target view (formed via the standard 3DGS projection of $\Sigma^{\text{world}}_g$ to image space; full derivation in App.~\ref{sec:method-app-opt}) and $\tau$ is a log-aspect budget. Operating on the screen-space projection aligns the penalty with the perceptual quantity of interest (whether the rendered Gaussian appears as a thin streak) and is well-defined for any per-Gaussian local rotation.

\paragraph{Background compositing and refinement objective.}
The target frames carry a foreground alpha mask. To avoid the asymmetric matte-boundary gradient of single-side compositing, we composite both rendered and target images against a shared per-iteration random RGB background $\mathbf{b}\!\sim\!\mathcal{U}([0,1]^3)$ before computing the loss; the resulting symmetric boundary gradient pulls the splatting alpha toward the matte without an explicit silhouette loss (full derivation in App.~\ref{sec:method-app-opt}). The refinement objective is
\begin{equation}
\mathcal{L}_{\text{ref}} = \lambda_1\,\mathcal{L}_1 + \lambda_{\text{ssim}}\,(1 - \text{SSIM}) + \mathbb{1}[t \geq 500] \cdot \lambda_{\text{aspect}}\,\mathcal{L}_{\text{aspect}}.
\label{eq:loss-ref}
\end{equation}
We omit LPIPS at this iteration budget; rationale and per-parameter learning rates are in App.~\ref{sec:method-app-opt}.

%% file: experiment.tex
\section{Experiments}
\label{sec:experiments}

\subsection{Datasets and Evaluation Metrics}
\label{sec:exp-bench}

\paragraph{Datasets.}
We train on CelebV-HQ~\citep{zhu2022celebvhq} (Phase~1) and
NeRSemble~v2~\citep{kirschstein2023nersemble} (Phase~2). For
feed-forward zero-shot evaluation we use VFHQ~\citep{xie2022vfhq}
with its default test split, and HDTF~\citep{zhang2021flow},
following the test split used in~\citep{chu2024gpavatar,chu2024generalizable},
including $19$ video clips. For per-subject evaluation we use the
SplattingAvatar~\citep{shao2024splattingavatar} dataset, with the
last $350$ frames of each video reserved as the test set. All
ablation tables (main and appendix) are evaluated on a held-out
CelebV-HQ~\citep{zhu2022celebvhq} test slice. VFHQ and HDTF do not
appear in any supervised stage of our training, so they constitute a
strict cross-domain zero-shot probe rather than an in-domain
benchmark.

\paragraph{Evaluation metrics.}
Standard image-quality metrics PSNR$\,\uparrow$, SSIM$\,\uparrow$, and
LPIPS$\,\downarrow$~\citep{zhang2018unreasonable} apply to both lanes.
For the feed-forward lane we additionally report ArcFace identity
similarity (CSIM$\,\uparrow$), average expression distance
(AED$\,\downarrow$), average pose distance (APD$\,\downarrow$), and
average keypoint distance (AKD$\,\downarrow$), all computed post-hoc
on existing renders. We evaluate two reenactment settings:
\emph{self-reenactment} sets the source to the first frame and the
remaining frames as driver/target; \emph{cross-reenactment} uses a
different-identity driver, has no pixel ground truth, and is
evaluated only by CSIM, AED, and APD. Refinement iteration count and
per-subject wall-clock are reported in Tab.~\ref{tab:efficiency}.

\subsection{Implementation Details}
\label{sec:exp-impl}

For Phase~1 we use CelebV-HQ for to train for $200$K iterations, and the encoder is initialized from DINOv3-B pretrain weight which is frozen during training. Phase~2 continues from the Phase-1 checkpoint on NeRSemble~v2 multiview data and a $25\%$ NeRSemble cross-time mix share. The per-subject refinement
runs the $10$K-iter training on given videos. All FLAME parameters are recovered with the EMICA-based
tracker~\citep{chu2024generalizable}. We use Adamw for both Phase~1 and Phase~2 at a base learning rate of $1{\times}10^{-4}$. Training
runs on a single H100~NVL GPU with an AMD~Ryzen Threadripper~3995X
host for 14 days.

\subsection{Main Results}
\label{sec:exp-main}

\begin{figure*}[t!]
\centering
\includegraphics[width=0.96\textwidth]{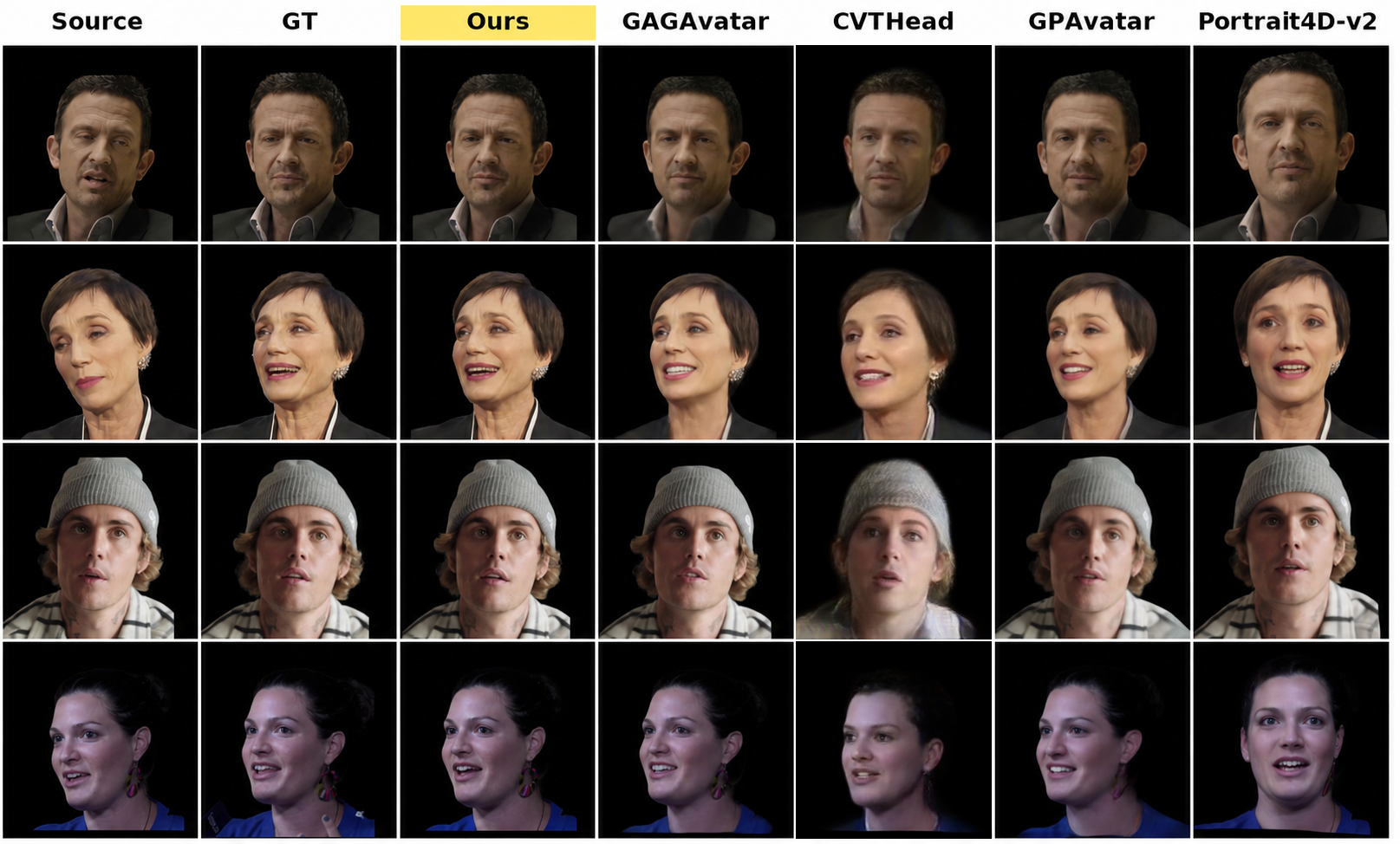}\\[2pt]
{\footnotesize (a) Self-reenactment samples from VFHQ}\\[6pt]
\includegraphics[width=0.96\textwidth]{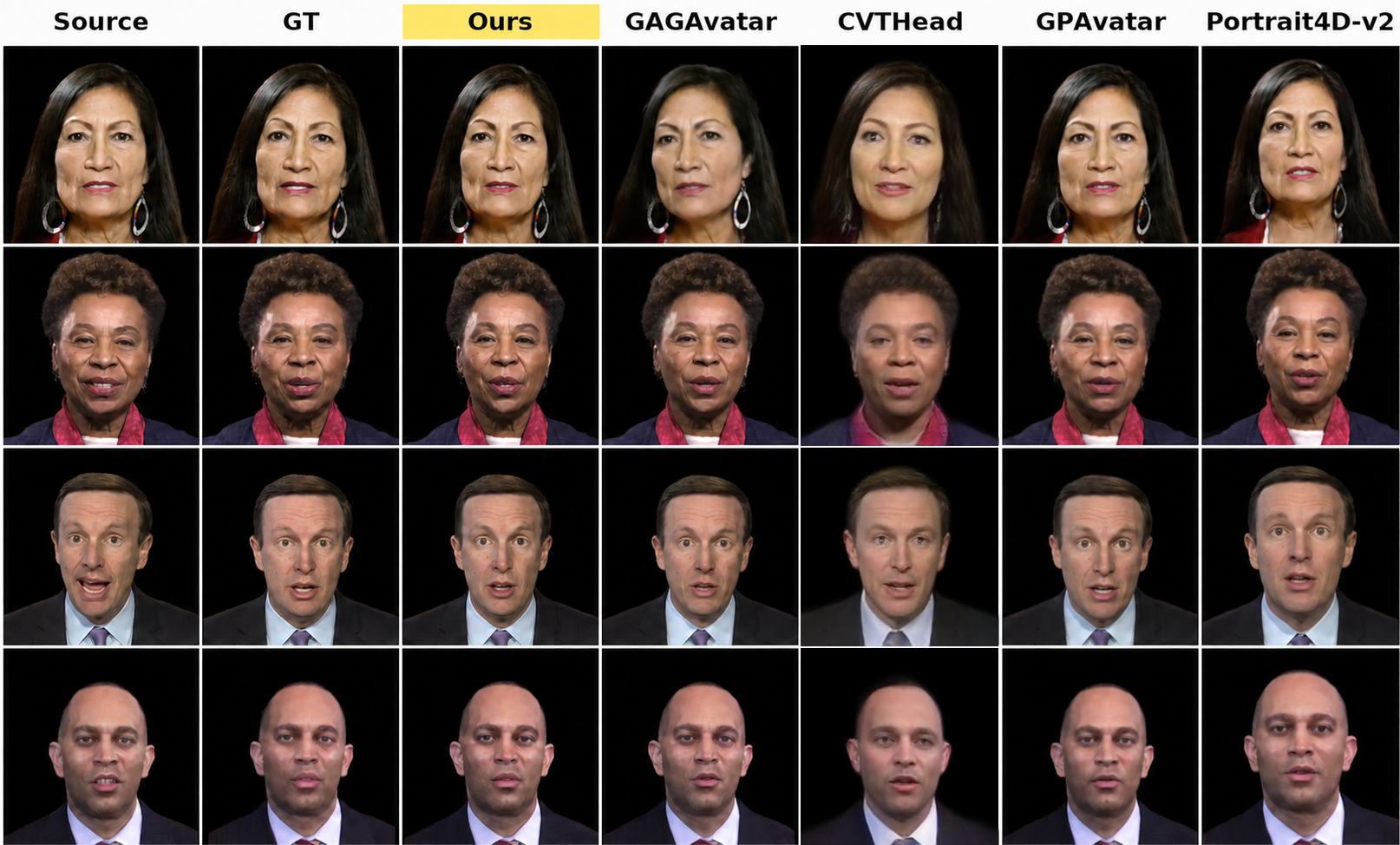}\\[2pt]
{\footnotesize (b) Self-reenactment samples from HDTF}
\caption{Feed-forward qualitative comparison on self-reenactment
(Tabs.~\ref{tab:main-ff-vfhq}--\ref{tab:main-ff-hdtf}).
The source is set to the first frame and the target to a held-out
frame from the same video. (a) Samples from VFHQ; (b) samples from
HDTF. We compare against GAGAvatar~\citep{chu2024generalizable},
CVTHead~\citep{ma2024cvthead}, GPAvatar~\citep{chu2024gpavatar},
and Portrait4D-v2~\citep{deng2024portrait4dv2}; our method is
highlighted in yellow.}
\label{fig:qual-cross-domain}
\vspace{-0.5cm}
\end{figure*}

\paragraph{Feed-forward comparison (Tabs.~\ref{tab:main-ff-vfhq}--\ref{tab:main-ff-hdtf}).}
We compare the feed-forward stage of our pipeline against eleven
generalizable head-avatar baselines on VFHQ and HDTF. VFHQ and HDTF
are out-of-distribution for our model (Phase~1 trains on
CelebV-HQ~\citep{zhu2022celebvhq}, Phase~2 on
NeRSemble~v2~\citep{kirschstein2023nersemble}), while several
baselines, notably GAGAvatar~\citep{chu2024generalizable} and
GPAvatar~\citep{chu2024gpavatar}, include one or both datasets in
their training distribution. Ours-FF leads every column on both
splits, surpassing the in-domain leader
GAGAvatar by ${+}1.5$\,dB PSNR on each, with consistent gains across
all reported metrics. Qualitative renders are in
Fig.~\ref{fig:qual-cross-domain}, and the cross-identity companion in
App.~\ref{sec:exp-app-cross-id}.

\begin{table}[t]
\centering
\caption{Feed-forward zero-shot reenactment on VFHQ. ``Self'' is
one-shot self-reenactment; ``Cross'' uses a different-identity driver
and reports only identity- and expression-similarity metrics. Best
per-column in \textbf{bold}.}
\label{tab:main-ff-vfhq}
\setlength{\tabcolsep}{3pt}
\resizebox{\columnwidth}{!}{%
\begin{tabular}{l ccccccc ccc}
\toprule
& \multicolumn{7}{c}{Self-reenactment} & \multicolumn{3}{c}{Cross-reenactment} \\
\cmidrule(lr){2-8}\cmidrule(lr){9-11}
Method
& PSNR\,$\uparrow$ & SSIM\,$\uparrow$ & LPIPS\,$\downarrow$
& CSIM\,$\uparrow$ & AED\,$\downarrow$ & APD\,$\downarrow$ & AKD\,$\downarrow$
& CSIM\,$\uparrow$ & AED\,$\downarrow$ & APD\,$\downarrow$ \\
\midrule
StyleHeat~\citep{yin2022styleheat}                 & 19.95 & 0.726 & 0.211 & 0.537 & 0.199 & 0.385 & 7.659 & 0.407 & 0.279 & 0.551 \\
ROME~\citep{khakhulin2022realistic}                & 19.96 & 0.786 & 0.192 & 0.701 & 0.138 & 0.186 & 4.986 & 0.530 & 0.259 & 0.277 \\
OTAvatar~\citep{ma2023otavatar}                    & 17.65 & 0.563 & 0.294 & 0.465 & 0.234 & 0.545 & 18.19 & 0.364 & 0.324 & 0.678 \\
HideNeRF~\citep{li2023hide}                        & 19.79 & 0.768 & 0.180 & 0.787 & 0.143 & 0.361 & 7.254 & 0.514 & 0.277 & 0.527 \\
CVTHead~\citep{ma2024cvthead}                      & 18.43 & 0.706 & 0.317 & 0.504 & 0.186 & 0.224 & 5.678 & 0.374 & 0.261 & 0.311 \\
GPAvatar~\citep{chu2024gpavatar}                   & 21.04 & 0.807 & 0.150 & 0.772 & 0.132 & 0.189 & 4.226 & 0.564 & 0.255 & 0.328 \\
Real3D-Portrait~\citep{ye2024real3dportrait}       & 20.88 & 0.780 & 0.154 & 0.801 & 0.150 & 0.268 & 5.971 & 0.663 & 0.296 & 0.411 \\
Portrait4D~\citep{deng2024portrait4d}              & 20.35 & 0.741 & 0.191 & 0.765 & 0.144 & 0.205 & 4.854 & 0.596 & 0.286 & 0.258 \\
Portrait4D-v2~\citep{deng2024portrait4dv2}         & 21.34 & 0.791 & 0.144 & 0.803 & 0.117 & 0.187 & 3.749 & 0.656 & 0.268 & 0.273 \\
GAGAvatar~\citep{chu2024generalizable}             & 21.83 & 0.818 & 0.122 & 0.816 & 0.111 & 0.135 & 3.349 & 0.633 & 0.253 & 0.247 \\
\textbf{Ours-FF}                                    & \textbf{23.34} & \textbf{0.875} & \textbf{0.077} & \textbf{0.916} & \textbf{0.091} & \textbf{0.061} & \textbf{2.181} & \textbf{0.675} & \textbf{0.245} & \textbf{0.210} \\
\bottomrule
\end{tabular}%
}
\end{table}

\begin{table}[t]
\centering
\caption{Feed-forward zero-shot reenactment on HDTF. Same column
structure as Tab.~\ref{tab:main-ff-vfhq}. Best per-column in
\textbf{bold}.}
\label{tab:main-ff-hdtf}
\setlength{\tabcolsep}{3pt}
\resizebox{\columnwidth}{!}{%
\begin{tabular}{l ccccccc ccc}
\toprule
& \multicolumn{7}{c}{Self-reenactment} & \multicolumn{3}{c}{Cross-reenactment} \\
\cmidrule(lr){2-8}\cmidrule(lr){9-11}
Method
& PSNR\,$\uparrow$ & SSIM\,$\uparrow$ & LPIPS\,$\downarrow$
& CSIM\,$\uparrow$ & AED\,$\downarrow$ & APD\,$\downarrow$ & AKD\,$\downarrow$
& CSIM\,$\uparrow$ & AED\,$\downarrow$ & APD\,$\downarrow$ \\
\midrule
StyleHeat~\citep{yin2022styleheat}                 & 21.41 & 0.785 & 0.155 & 0.657 & 0.158 & 0.162 & 4.585 & 0.632 & 0.271 & 0.239 \\
ROME~\citep{khakhulin2022realistic}                & 20.51 & 0.803 & 0.145 & 0.738 & 0.133 & 0.123 & 4.763 & 0.726 & 0.268 & 0.191 \\
OTAvatar~\citep{ma2023otavatar}                    & 20.52 & 0.696 & 0.166 & 0.662 & 0.180 & 0.170 & 8.295 & 0.643 & 0.292 & 0.222 \\
HideNeRF~\citep{li2023hide}                        & 21.08 & 0.811 & 0.117 & 0.858 & 0.120 & 0.247 & 5.837 & 0.843 & 0.276 & 0.288 \\
CVTHead~\citep{ma2024cvthead}                      & 20.08 & 0.762 & 0.179 & 0.608 & 0.169 & 0.138 & 4.585 & 0.591 & 0.242 & 0.203 \\
GPAvatar~\citep{chu2024gpavatar}                   & 23.06 & 0.855 & 0.104 & 0.855 & 0.114 & 0.135 & 3.293 & 0.842 & 0.268 & 0.219 \\
Real3D-Portrait~\citep{ye2024real3dportrait}       & 22.82 & 0.835 & 0.103 & 0.851 & 0.138 & 0.137 & 4.640 & 0.903 & 0.299 & 0.238 \\
Portrait4D~\citep{deng2024portrait4d}              & 20.81 & 0.786 & 0.137 & 0.810 & 0.134 & 0.131 & 4.151 & 0.793 & 0.291 & 0.240 \\
Portrait4D-v2~\citep{deng2024portrait4dv2}         & 22.87 & 0.860 & 0.105 & 0.860 & 0.111 & 0.111 & 3.292 & 0.857 & 0.262 & 0.183 \\
GAGAvatar~\citep{chu2024generalizable}             & 23.13 & 0.863 & 0.103 & 0.862 & 0.110 & 0.111 & 2.985 & 0.851 & 0.231 & 0.181 \\
\textbf{Ours-FF}                                    & \textbf{24.67} & \textbf{0.916} & \textbf{0.076} & \textbf{0.871} & \textbf{0.103} & \textbf{0.078} & \textbf{2.326} & \textbf{0.905} & \textbf{0.224} & \textbf{0.147} \\
\bottomrule
\end{tabular}%
}
\end{table}

Ours-FF preserves high-frequency identity-discriminative cues such
as forehead wrinkles, malar elevation, beard texture, nasolabial
folds, and hair-edge silhouette. The detail preservation is
consistent across identities and poses.

\begin{wraptable}{r}{0.55\textwidth}
\centering
\setlength{\tabcolsep}{4pt}
\caption{Per-subject training efficiency on a single RTX~3090.
Inference FPS at $512^{2}$ via the $3$-channel
\texttt{diff\_gaussian\_rasterization} backend
($58{,}173$~Gaussians/subject).}
\label{tab:efficiency}
\resizebox{0.55\textwidth}{!}{%
\begin{tabular}{l c c c c}
\toprule
Method & Iters\,$\downarrow$ & Train\,$\downarrow$ & Lat.\,(ms)\,$\downarrow$ & FPS\,$\uparrow$ \\
\midrule
FlashAvatar~\citep{xiang2024flashavatar}                & $150$K       & ${\sim}1.66$\,h           & --            & 291.20 \\
GaussianAvatars~\citep{qian2024gaussianavatars}         & $600$K       & ${\sim}9.25$\,h           & --            & 19.11  \\
GeoAvatar~\citep{moon2025geoavatar}                     & $300$K       & ${\sim}4.90$\,h           & --            & 71.52  \\
\midrule
\textbf{Ours+S3}                                        & \textbf{$10$K} & \textbf{${\sim}2.0$\,min} & \textbf{3.77} & \textbf{265.60}  \\
\bottomrule
\end{tabular}%
}
\end{wraptable}
\paragraph{Per-subject comparison (Tab.~\ref{tab:main-persubj}).}
We compare against per-subject baselines on the
SplattingAvatar~\citep{shao2024splattingavatar} dataset, with
GAGAvatar's one-shot result included in the leftmost column as a
feed-forward upper bound. Ours+S3 leads every reported metric (MSE,
PSNR, SSIM, LPIPS), surpassing the prior PSNR/SSIM/MSE leader
GeoAvatar~\citep{moon2025geoavatar} by $\mathbf{{+}1.3}$\,dB PSNR at
a $30{\times}$ shorter iteration budget ($10$K vs $300$K) and a
substantially shorter wall-clock (Tab.~\ref{tab:efficiency}).
Qualitative renders are in App.~\ref{sec:exp-app-persubj-qual},
Fig.~\ref{fig:qual-persubj}.

\begin{table}[t]
\centering
\caption{Per-subject monocular reenactment on the SplattingAvatar
dataset. The first column reports GAGAvatar's one-shot result as a
feed-forward reference. Best per-row in \textbf{bold}.}
\label{tab:main-persubj}
\setlength{\tabcolsep}{4pt}
\resizebox{\columnwidth}{!}{%
\begin{tabular}{l|cccccccccc}
\toprule
Metric
& \scriptsize GAGAv$^*$
& \scriptsize INSTA
& \scriptsize 3DGS
& \scriptsize SpA
& \scriptsize MonoGA
& \scriptsize FlashAv
& \scriptsize GA
& \scriptsize GA$_0$
& \scriptsize GeoAv
& \scriptsize \textbf{Ours+S3} \\
\midrule
MSE\,$\!\!\times\!\!10^{-3}\,\downarrow$
& 4.425 & 1.555 & 1.761 & 2.929 & 1.314 & 1.173 & 1.223 & 1.075 & 0.545 & \textbf{0.402} \\
PSNR\,$\uparrow$
& 23.541 & 28.083 & 27.543 & 25.333 & 28.813 & 29.306 & 29.124 & 29.686 & 32.635 & \textbf{33.960} \\
SSIM\,$\uparrow$
& 0.875 & 0.938 & 0.923 & 0.933 & 0.937 & 0.943 & 0.938 & 0.934 & 0.965 & \textbf{0.971} \\
LPIPS\,$\!\!\times\!\!10^{-1}\,\downarrow$
& 1.121 & 0.678 & 1.019 & 0.588 & 0.733 & 0.444 & 0.494 & 0.529 & 0.367 & \textbf{0.355} \\
\bottomrule
\end{tabular}%
}
\par\smallskip
\footnotesize
$^*$one-shot feed-forward reference (no per-subject training); SpA =
SplattingAvatar, MonoGA = MonoGaussianAvatar, GA = GaussianAvatars,
GeoAv = GeoAvatar. GaussianAvatars$_0$ is GaussianAvatars with SH degree set
to $0$ (per~\citep{moon2025geoavatar}~\S4.4). The Ours-FF feed-forward column
on this benchmark is omitted from the main paper for compactness.

\end{table}

\paragraph{Per-subject creation efficiency (Tab.~\ref{tab:efficiency}).}

We compare per-subject training cost on the SplattingAvatar set
against the per-subject baselines on a single RTX~3090.
Tab.~\ref{tab:efficiency} shows our per-subject refinement completes
in ${\sim}2$\,min per subject, two orders of magnitude shorter than
FlashAvatar (${\sim}1.66$\,h), GeoAvatar (${\sim}4.90$\,h), and
GaussianAvatars (${\sim}9.25$\,h) reported on RTX~3090. Our
$10$K-iter budget is $15{\times}$, $30{\times}$, and $60{\times}$
shorter than the per-subject schedules of FlashAvatar, GeoAvatar,
and GaussianAvatars respectively. Inference renders at
$265.6$\,FPS at $512^{2}$ on a single RTX~3090, real-time and within
the same throughput regime as the per-subject baselines; the
feed-forward stage produces a complete avatar from a single image in
${\sim}50$\,ms on the same RTX~3090.

\subsection{Ablation Study}
\label{sec:exp-ablations}

We report the headline pipeline-and-architecture ablation here on
self-reenactment over a held-out CelebV-HQ test slice. Remaining
ablations are deferred to App.~\ref{sec:exp-app-extras} and
App.~\ref{sec:method-app-training}.

\paragraph{Pipeline and architecture (Tab.~\ref{tab:ablation}).}
The top block (rows~A--D) builds the pipeline cumulatively, and the
bottom block (rows~E--G) holds (D) fixed and removes one
architectural component at a time.
In (A) we report Phase~1 alone, our feed-forward generator trained
only on monocular CelebV-HQ. Adding the Phase-2 NeRSemble~v2
post-training without controls (B) catastrophically forgets the
CelebV-HQ distribution. Reintroducing the L2-SP anchor at
$\lambda{=}10^{-3}$ together with the $25\%$ NeRSemble cross-time mix
share (C) recovers in-domain quality at near-equal NeRSemble quality,
indicating that the two controls are jointly sufficient. Adding the
per-subject refinement on top (D) yields the largest single jump
and we adopt (D) as our \textbf{Full} configuration throughout the
paper.
(E) removes the FLAME-conditioned residual head and regresses on all
metrics, identifying it as broadly load-bearing.
(F) removes the Hutchinson Jacobian penalty, which constrains the
position and scale heads to be view-invariant: geometry should be
an intrinsic property of the scene rather than depend on viewing
direction. Without this constraint, AKD rises substantially ($1.87$ vs $1.37$). (G) replaces $K$-source mean-pooling with attention-pooling, winning
SSIM (tied) and APD by small margins but losing on other metrics.
The three failure modes target distinct axes of the pipeline and support that the components are not redundant.

\begin{table}[t]
\centering
\caption{Pipeline and architecture ablation on a held-out CelebV-HQ
test slice. Top block (rows~A--D): cumulative stage-by-stage build-up;
row~D is our \textbf{Full} configuration. Bottom block: subtractive
ablations of architectural components, evaluated on top of row~D's
configuration.}
\label{tab:ablation}
\setlength{\tabcolsep}{4pt}
\resizebox{\columnwidth}{!}{%
\begin{tabular}{c l ccccccc}
\toprule
& Experiment
& PSNR\,$\uparrow$ & SSIM\,$\uparrow$ & LPIPS\,$\downarrow$
& CSIM\,$\uparrow$ & AED\,$\downarrow$ & APD\,$\downarrow$ & AKD\,$\downarrow$ \\
\midrule
(A) & Phase 1 only                                              & 24.84 & 0.879 & 0.073 & 0.923 & 0.086 & 0.053 & 2.102 \\
\rowcolor{yellow!18}
(B) & (A) + Phase 2 (no anchor, no cross-time mix)              & 22.73 & 0.799 & 0.135 & 0.814 & 0.117 & 0.112 & 3.244 \\
\rowcolor{green!10}
(C) & (B) + L2-SP anchor + 25\% cross-time mix                  & 24.94 & 0.881 & 0.070 & 0.935 & 0.084 & 0.054 & 2.032 \\
\rowcolor{green!22}
(D) & (C) + per-subject refinement (10K iters, \textbf{Full})   & \textbf{29.95} & 0.947 & 0.045 & \textbf{0.959} & \textbf{0.057} & 0.039 & \textbf{1.365} \\
\midrule
\rowcolor{gray!10}
(E) & (D) w/o residual (delta) head                             & 28.36 & 0.933 & 0.060 & 0.931 & 0.081 & 0.041 & 1.412 \\
\rowcolor{gray!10}
(F) & (D) w/o Jacobian penalty ($\lambda_{\text{jac}}{=}0$)     & 28.72 & 0.937 & \textbf{0.040} & 0.952 & 0.061 & 0.042 & 1.874 \\
\rowcolor{gray!10}
(G) & (D) attention pool (replaces mean-pool)                   & 29.46 & \textbf{0.948} & 0.047 & 0.907 & 0.062 & \textbf{0.036} & 1.374 \\
\bottomrule
\end{tabular}%
}
\vspace{-0.5cm}
\end{table}

%% file: method_appendix.tex
\section{Additional method details}
\label{sec:method-appendix}

This appendix collects the derivations, exact functional forms, and training-recipe specifics referenced from Section~\ref{sec:method}.

\subsection{Feed-forward representation and architecture}
\label{sec:method-app-arch}

\paragraph{UV-aligned representation.}
The UV-aligned per-pixel representation rests on three properties. First, the UV unwrap is topology-respecting: neighboring UV pixels correspond to neighboring mesh triangles, so a 2D convolutional generator inherits the inductive bias for face-bound primitives, and smoothness in the UV domain corresponds directly to smoothness on the mesh surface. Second, the Gaussian count is determined by the UV grid topology and is independent of the input image content; alternative representations must instead supply this information through explicit prediction or densification heuristics. Third, the UV alignment yields a spatially-local mapping from image features to per-Gaussian attributes: each UV pixel governs a single Gaussian, and no cross-pixel attention or order-agnostic pooling is required. The architecture and conditioning interface are identical across the two training phases; only the data distribution differs.

\paragraph{Barycentric feature warp.}
The UV feature map is constructed by a parameter-free barycentric warp from the encoder output. For each UV pixel, we identify its FLAME triangle from a precomputed map, interpolate the corresponding 3D position from the source-frame mesh vertices using the triangle's barycentric weights, project the 3D point through the source camera, and bilinearly sample the encoder feature and the source RGB at the projected location.

\paragraph{Harmonic view-direction encoding.}
The target camera direction $\mathbf{d}$ is the unit-normalized third column of the target view's world-to-camera matrix, encoded as the $27$-dimensional harmonic embedding~\citep{mildenhall2020nerf}
\begin{equation}
\gamma(\mathbf{d}) = \big[\mathbf{d},\; \sin(2^{k}\pi\mathbf{d}),\; \cos(2^{k}\pi\mathbf{d})\big]_{k=0..3} \in \mathbb{R}^{27}.
\end{equation}

\paragraph{FLAME version, tracker, and UV layout.} Source and target FLAME parameters are produced by the GAGAvatar tracker stack~\citep{chu2024generalizable} (\texttt{xg-chu/GAGAvatar\_track}; an EMICA-style monocular tracker built on the EMOCA / SMIRK expression-aware lineage, combined with a VGGHead detector, StyleMatte matting, and a 300-step Adam OptimEngine). Per frame, the tracker emits $\boldsymbol{\beta}\in\mathbb{R}^{300}$ (shape), $\boldsymbol{\psi}\in\mathbb{R}^{100}$ (\textsc{expcode}, FLAME 2020 expression blendshape coefficients), and a tracker-specific subset of FLAME pose: $\boldsymbol{\theta}^{\text{pose}}\in\mathbb{R}^{6}$ (\textsc{posecode}, $=$ global head rotation $\oplus$ jaw axis-angle) and $\boldsymbol{\theta}^{\text{eye}}\in\mathbb{R}^{6}$ (\textsc{eyecode}, bilateral eye pose). The standard FLAME pose vector~\citep{li2017learning} is $\theta\in\mathbb{R}^{3K+3}=\mathbb{R}^{15}$ for $K\!=\!4$ joints (neck, jaw, two eyeballs) plus global rotation; the EMICA tracker absorbs the neck joint into the camera transform and exposes the eye pose separately, yielding the 6+6 subset above. The tracker also emits a $3\!\times\!4$ camera transform and a face bounding box. The mesh is FLAME 2020~\citep{li2017learning} extended via \texttt{patch\_teeth.py} to $5143$ vertices ($10144$ faces), unwrapped onto a $256\!\times\!256$ UV grid via \texttt{flame\_uv.npz} ($5118$ UV vertices, $9976$ UV faces, no UV coordinates for teeth); the teeth-patched mesh and the UV unwrap are released as data assets of the GAGAvatar tracker~\citep{chu2024generalizable}. The valid-UV mask $\Omega \subset \{1,\dots,256\}^2$ has $|\Omega|\!\approx\!58{,}173$ pixels, fixed across all subjects and frames; this is the count of Gaussians per identity.

\paragraph{Residual head conditioning vector.} The residual head consumes a $112$-dimensional conditioning vector formed by concatenating the target frame's expression, pose, and eye codes:
\begin{equation}
\mathbf{c}_t \;=\; \boldsymbol{\psi}_t \,\oplus\, \boldsymbol{\theta}^{\text{pose}}_t \,\oplus\, \boldsymbol{\theta}^{\text{eye}}_t \;\in\; \mathbb{R}^{100+6+6} = \mathbb{R}^{112}.
\end{equation}
A 2-layer MLP produces per-attribute FiLM~\citep{perez2018film} coefficients $(\boldsymbol{\gamma}_a, \boldsymbol{\beta}_a)$ for $a\in\{\text{position},\text{rotation},\text{scale},\text{color}\}$; the modulated features pass through a $1\!\times\!1$ convolutional head whose output is added to the post-decoder UV maps for those four attributes. Opacity is excluded (we observed alpha-pumping during animation in early experiments, an empirical observation rather than a derived property). Shape $\boldsymbol{\beta}$ is intentionally \emph{excluded} from $\mathbf{c}_t$: shape is fixed-per-subject, feeds mesh geometry directly through FLAME's blendshape basis (Eq.~\ref{eq:face-binding}), and is the load-bearing axis of cross-subject generalization; conditioning the residual on $\boldsymbol{\beta}$ would let the head learn per-identity displacements and overfit to training subjects rather than identity-agnostic expression-conditioned residuals. Head pose $\boldsymbol{\theta}^{\text{pose}}_{[0:3]}$ is included in $\mathbf{c}_t$ but does not feed the harmonic view encoder $\gamma(\mathbf{d})$, so the residual head is structurally independent of view direction (App.~\ref{sec:method-app-jacobian}). The residual head's contribution is evaluated in our ablation study (Tab.~\ref{tab:ablation}).

\paragraph{Per-triangle frame construction.}
The per-triangle local frame $(\mathbf{c}_f, \mathbf{R}_f, s_f)$ for a triangle with vertices $(\mathbf{v}_0, \mathbf{v}_1, \mathbf{v}_2)$ is constructed as follows. The center $\mathbf{c}_f$ is the triangle centroid. The rotation $\mathbf{R}_f$ is built from the edge vectors via Gram-Schmidt-like orthonormalization: $\mathbf{a}_0 = (\mathbf{v}_1 - \mathbf{v}_0)/\|\mathbf{v}_1 - \mathbf{v}_0\|$ along the first edge, $\mathbf{a}_1$ along the unit triangle normal, and $\mathbf{a}_2 = \mathbf{a}_1 \times \mathbf{a}_0$. The isotropic scale is the average of one edge length and one altitude. This averaging follows the textual description in GaussianAvatars~\citep{qian2024gaussianavatars} \S3.2 (``the mean length of one of the edges and its perpendicular'') and matches the public reference implementation; the resulting scalar inherits the FLAME mesh's fixed vertex ordering, and the thin-triangle fallback below bounds the residual conditioning:
\begin{equation}
s_f = \tfrac{1}{2}\!\left(\|\mathbf{v}_1 - \mathbf{v}_0\| + |\mathbf{a}_2 \cdot (\mathbf{v}_2 - \mathbf{v}_0)|\right).
\end{equation}
All $\mathbf{R}_f$ and $s_f$ are computed with respect to the fixed FLAME 2020 vertex ordering ($5143$ vertices via \texttt{patch\_teeth.py}); changing the mesh re-permutes vertex indices and invalidates the predicted local quaternions $\mathbf{q}^{\text{local}}_g$, since the residual head's quaternion correction is interpreted in this fixed local basis.

\paragraph{Numerical stability on degenerate triangles.}
The FLAME UV unwrap contains a small number of near-degenerate triangles at eye corners, lip seams, and ear seams. We rely on safe normalization in the Gram-Schmidt construction (rather than an explicit detection-and-fallback branch) to prevent division-by-zero on near-zero edge or altitude lengths; the per-triangle scale $s_f$ tends to $\|\mathbf{v}_1 - \mathbf{v}_0\|/2$ in the zero-altitude limit and remains finite. Empirically the residual conditioning has not driven divergent training or visible artifacts.
\subsection{Training procedure}
\label{sec:method-app-training}

\paragraph{Data-difficulty-weighted source-frame sampler.}
We bias the source distribution against trivial near-identity source-target pairs by sampling source$[0]$ from a softmax over a weighted combination of camera-direction and expression-code dissimilarity from the target frame. The softmax temperature is annealed monotonically during the early phase of training: at high temperature the softmax is near-uniform and the sampler draws candidates approximately uniformly across the pool, while at low temperature the sampler concentrates on candidates with maximum dissimilarity. The temperature is annealed from $\tau_0\!=\!2.0$ (near-uniform) to $\tau_T\!=\!0.5$ (concentrated on maximum-dissimilarity candidates) over the first quarter of training via a piecewise-linear schedule; only \texttt{source[0]} is drawn from this softmax, the remaining $K\!-\!1$ sources are uniform.
\paragraph{Phase 2 specifics.}
Each Phase 2 batch samples $K\!+\!1$ cameras from the synchronized 16-camera NeRSemble capture, mixing cross-camera draws (same identity, same instant, $K\!+\!1$ distinct cameras; default $75\%$ of batches) with cross-time draws (same identity, $K\!+\!1$ distinct time-shifted views from the available cameras, reducing to monocular same-camera-different-time when $K\!=\!1$; default $25\%$). The cross-camera majority provides multi-view supervision absent from Phase 1; the cross-time minority preserves the wide-identity prior acquired during Phase 1 and prevents collapse onto the smaller multi-view training set. Phase 2 uses layer-wise learning-rate decay following the LCA Appendix~C scheme~\citep{li2026lca} (we denote this rate $\gamma_{\text{LR}}\!=\!0.65$ to avoid notational collision with $\gamma(\mathbf{d})$) on the encoder layers, with normalization-layer running statistics continued (not reset, not frozen) across the Phase 1 → Phase 2 boundary, with both decoders held at base LR, supplemented by an L2-SP anchor~\citep{li2018l2sp} against the Phase 1 checkpoint as a regularization-to-Phase-1-prior. The motivating concern is identity-prior collapse onto the 17 NeRSemble subjects~\citep{kirschstein2023nersemble} we train on (a subset of the released 220-subject pool) relative to CelebV-HQ's wide-identity Phase-1 manifold ($\sim\!15{,}653$ identities); our ablation study (Tab.~\ref{tab:ablation}) reports whether without-anchor exhibits regression on CelebV-HQ-held-out subjects. The cross-cam-mix ratio, LR-decay coefficient, and L2-SP weight are not claimed to be optimal. Remaining optimizer, batch-size, and hardware details are in the ``Optimizer, precision, and batch sizes'' paragraph below.

\paragraph{Loss-form table.} Let $\Omega\subset\{1,\dots,256\}^2$ denote the valid-UV mask ($|\Omega|\!\approx\!58{,}173$). The seven currently-undefined loss terms in Eq.~\ref{eq:loss-ff} have explicit forms:
\begin{align*}
\mathcal{L}_1 &= \tfrac{1}{HW}\big\|\hat{\mathbf{I}}-\mathbf{I}\big\|_1, \\
\mathcal{L}_{\text{ms}} &= 1 - \mathrm{MS\text{-}SSIM}_{\sigma\in\{0.5,1,2,4,8\}}(\hat{\mathbf{I}}, \mathbf{I}), \\
\mathcal{L}_{\text{lp}} &= \mathrm{LPIPS\text{-}AlexNet}(\hat{\mathbf{I}}, \mathbf{I}), \\
\mathcal{L}_{\text{box}} &= \tfrac{1}{|\text{bbox}|}\big\|\hat{\mathbf{I}}_{\text{bbox}}-\mathbf{I}_{\text{bbox}}\big\|_1,\quad\text{bbox: EMICA face bbox dilated} \times 1.65, \\
\mathcal{L}_{\text{ps}} &= \tfrac{1}{|\Omega|}\sum_{(u,v)\in\Omega}\big(\big\|\nabla_u \mathbf{x}^{\text{local}}\big\|_1 + \big\|\nabla_v \mathbf{x}^{\text{local}}\big\|_1\big), \\
\mathcal{L}_{\text{ss}} &= \tfrac{1}{|\Omega|}\sum_{(u,v)\in\Omega}\big(\big\|\nabla_u \boldsymbol{s}^{\log}\big\|_1 + \big\|\nabla_v \boldsymbol{s}^{\log}\big\|_1\big), \\
\mathcal{L}_\delta &= \tfrac{1}{|\Omega|}\sum_{a \in \{p,r,s,c\}} \big\|\Delta_a^{\Omega}\big\|_1\quad\text{(omits opacity)}, \\
\mathcal{L}_{\text{jac}} &\text{ as in body Eq.~\ref{eq:loss-jac}}.
\end{align*}
The reduction is mean over valid pixels for all UV-domain terms. The TV penalties ($\mathcal{L}_{\text{ps}}$, $\mathcal{L}_{\text{ss}}$) are defined on the UV grid; at FLAME UV seams (face$\leftrightarrow$ear, neck) UV-adjacent pixels are not mesh-adjacent, and the TV imposes an unintended smoothing across the seam. The cross-seam contribution to $|\Omega|$ is small in our setting. Order-of-magnitude weights are $\lambda_1, \lambda_{\text{ms}}, \lambda_{\text{lp}}, \lambda_{\text{box}}\!\sim\!10^0$, $\lambda_{\text{ps}}, \lambda_{\text{ss}}\!\sim\!10^{-2}$, $\lambda_{\text{jac}}\!\sim\!10^{-1}$, and $\lambda_\delta\!\sim\!10^{-2}$ in Phase 2 (the residual-shrinkage term is inactive in Phase 1, where the head trains under wide-identity exposure with weight decay only). The two-phase contribution, the L2-SP anchor, and cross-domain CelebV-HQ behavior are reported in our ablation study (Tab.~\ref{tab:ablation}).

\paragraph{Optimizer, precision, and batch sizes.}
Phase~1 uses plain Adam at $2.5{\times}10^{-4}$ with FP32 precision,
batch size $48$ on a single H100~NVL, and gradient clipping at norm
$5.0$. Phase~2 uses AdamW with cosine warmup, bf16 mixed precision,
and batch size $12$; the L2-SP weight is
$\lambda_{\text{sp}}{=}10^{-3}$ and the NeRSemble cross-time mix share
is $25\%$ (the layer-wise LR decay $\gamma_{\text{LR}}$, anchor
mechanism, and the cross-camera majority are described in Phase~2
specifics above). The
residual-shrinkage term $\mathcal{L}_\delta$ is inactive in Phase~1 and
activated in Phase~2; the remaining photometric loss weights match the
Phase-1 magnitudes (Loss-form table above). The per-subject refinement
sets $\lambda_{\text{aspect}}{=}0.5$ for the screen-space anti-anisotropy
penalty active over iterations $500$--$10{,}000$; per-parameter
learning rates and the global cosine schedule are described in
App.~\ref{sec:method-app-opt}.

\subsection{Jacobian penalty derivation}
\label{sec:method-app-jacobian}

\paragraph{Goal: per-pixel view-direction invariance.}
Let $f_{\text{geo}}$ denote the position and scale heads. We require the per-pixel Jacobian $J_p = \partial f_{\text{geo},p}/\partial \gamma(\mathbf{d}) \in \mathbb{R}^{d_{\text{out}} \times 27}$ to vanish for every UV pixel $p \in \Omega$. The natural penalty is the per-pixel Frobenius energy
\begin{equation}
\mathcal{F} \;=\; \tfrac{1}{|\Omega|}\sum_{p \in \Omega} \|J_p\|_F^2 \;=\; \tfrac{1}{|\Omega|}\sum_{p}\sum_{i,j} J_{p,i,j}^2.
\label{eq:jac-frobenius-true}
\end{equation}
A direct evaluation of $\mathcal{F}$ requires either $|\Omega| \approx 58$K backward passes against $\gamma(\mathbf{d})$, or a \texttt{vmap-jacrev} materialization of the $|\Omega| \times d_{\text{out}} \times 27$ per-pixel Jacobian tensor; both inflate training-time autograd cost by one to two orders of magnitude.

\paragraph{Discrete two-point alternative.}
A finite-difference proxy
\begin{equation}
\big\|f_{\text{geo}}(\mathbf{x}_{\text{src}}, \mathbf{d}) - f_{\text{geo}}(\mathbf{x}_{\text{src}}, \mathbf{d}^{\text{ref}})\big\|_1
\end{equation}
between the actual view direction $\mathbf{d}$ and a fixed reference $\mathbf{d}^{\text{ref}}$ avoids autograd cost but, at the moderate-to-large angular separations typical of our training videos, confounds first-order view-direction sensitivity with higher-order curvature; we prefer the differential limit below.

\paragraph{Hutchinson per-pixel estimator.}
Let $f_{\text{geo}, p} \in \mathbb{R}^{d_{\text{out}}}$ denote the position or scale prediction at UV pixel $p$, with per-pixel Jacobian $J_p = \partial f_{\text{geo},p}/\partial \gamma(\mathbf{d}) \in \mathbb{R}^{d_{\text{out}} \times 27}$ ($d_{\text{out}}\!=\!3$ for both heads). Sample i.i.d.\ Rademacher $\mathbf{v}_p \in \{-1, +1\}^{d_{\text{out}}}$ and define $g(\gamma) = \sum_p \langle \mathbf{v}_p, f_{\text{geo}, p}(\gamma) \rangle$. Then
\begin{equation}
\mathbb{E}_{\mathbf{v}}\big[\big\|\nabla_\gamma g\big\|_2^2\big] \;=\; \sum_p \big\|J_p\big\|_F^2.
\label{eq:hutchinson-identity}
\end{equation}
\emph{Proof.} $\nabla_\gamma g = \sum_p J_p^\top \mathbf{v}_p$; by independence and zero mean, $\mathbb{E}[\mathbf{v}_p \mathbf{v}_q^\top] = \delta_{pq}\mathbf{I}$, so $\mathbb{E}[\|\nabla_\gamma g\|_2^2] = \sum_{p,q} \mathbb{E}[\mathbf{v}_p^\top J_p J_q^\top \mathbf{v}_q] = \sum_p \mathrm{tr}(J_p J_p^\top) = \sum_p \|J_p\|_F^2$.~$\square$

\paragraph{Estimator variance and bias.}
The lineage anchor is Hutchinson~\citep{hutchinson1989stochastic} (Rademacher trace estimator); the single-backward Jacobian-regularization scheme follows Hoffman et al.~\citep{hoffman2019robust} (whose official implementation uses Gaussian-unit-norm projections, agreeing with Rademacher in expectation up to a $d_{\text{out}}$ factor, with comparable but non-identical per-sample variance); per-feature variants are due to Drucker and LeCun~\citep{drucker1992improving} and Bishop~\citep{bishop1995regularization}. We use Rademacher per Hutchinson 1989, with $d_{\text{out}}\!=\!3$ for both position and scale heads applied independently. In the body penalty (Eq.~\ref{eq:loss-jac}), $g_{\mathbf{x}}\!=\!\sum_p \langle \mathbf{v}^{\mathbf{x}}_p, \mathbf{x}^{\text{local}}_p\rangle$ and $g_{\boldsymbol{s}}\!=\!\sum_p \langle \mathbf{v}^{\boldsymbol{s}}_p, \boldsymbol{s}^{\text{local}}_p\rangle$ are the per-head Hutchinson scalars, with independently sampled Rademacher projections.

\textbf{Estimator-target gap.} The Hutchinson identity (Eq.~\ref{eq:hutchinson-identity}) shows $\mathbb{E}_{\mathbf{v}}[\|\nabla_\gamma g\|_2^2] = \sum_p\|J_p\|_F^2$ on the inner argument; the wrapped loss is biased relative to $\log(1+\sum_p\|J_p\|_F^2)$ by Jensen's inequality (strict whenever $\mathrm{Var}[\|\nabla_\gamma g\|^2] > 0$). By Taylor expansion of $\log(1+\cdot)$ around $\mu = \mathbb{E}[\|\nabla_\gamma g\|^2]$, the wrapped-loss bias scales as $-\frac{1}{2}\mathrm{Var}[\|\nabla_\gamma g\|^2] / (1+\mu)^2$ at leading order; the bias decays toward zero as training drives $\mu$ toward zero. The penalty therefore targets the per-pixel Frobenius energy via a Hutchinson estimate of the inner argument, with a second-order bias from the $\log(1+\cdot)$ wrapping. The leading-order Taylor expansion above is per-head; the wrapped loss applies a single $\log(1+\cdot)$ over the sum of the position and scale contributions, so the joint Jensen bias is governed by $\mathrm{Var}[\|\nabla_\gamma g_{\mathbf{x}}\|^2] + \mathrm{Var}[\|\nabla_\gamma g_{\boldsymbol{s}}\|^2]$ rather than the sum of two independently-wrapped per-head $\log$ biases.

\begin{sloppypar}
\textbf{Per-iteration variance.} The relative variance of the inner argument scales as $\mathcal{O}(\sum_{p\neq q}\|J_p J_q^\top\|_F^2 / (\sum_p\|J_p\|_F^2)^2) = \Theta(1)$ in $|\Omega|$ (the cross-pixel sum is $\Theta(|\Omega|^2)$, the squared-Frobenius normalizer is also $\Theta(|\Omega|^2)$, leading-order in $|\Omega|$, with diagonal contribution $\Theta(|\Omega|)$ dominated for $|\Omega| \gg 1$). The estimator is therefore not asymptotically variance-shrinking with pixel count; the regularizer contributes a controlled-variance inductive bias whose per-step gradient is dominated by the photometric trio at the chosen $\lambda_{\text{jac}} \sim 10^{-1}$ (consistent with the magnitude in the loss-form table above). The $\Theta(1)$ relative-variance scaling above holds under approximate cross-pixel independence; correlated $J_p$ across UV pixels (e.g., overlapping CNN receptive fields) would re-introduce a cross-pixel scaling factor. We verify training stability empirically.\end{sloppypar}

\textbf{Why Rademacher per-pixel and not unprojected pixel-sum.} Per the Hutchinson identity (Eq.~\ref{eq:hutchinson-identity}), the per-pixel Rademacher projection makes the cross-pixel cancellation regime independent of any sign correlation in $\{J_p\}$, regardless of training equilibrium: an unprojected pixel-sum $\big\|\sum_p \nabla_\gamma f_p\big\|_2^2$ would admit a benign minimizer at sign-cancelled $\{J_p\}$; the per-pixel projection removes this failure mode by construction. The penalty matches the autograd cost of an unprojected pixel-sum (one backward pass against $\gamma(\mathbf{d})$).

\paragraph{Why we wrap in $\log(1+\cdot)$.}
We wrap the inner argument in $\log(1+\cdot)$ to cap the regularizer's per-step magnitude during early training: at random initialization the unwrapped argument is on the order of $10^{2}$ and would otherwise dominate the photometric signal. At convergence the wrapper behaves as the identity (gradient $\to 1/(1+\|\cdot\|^2)$, vanishing as the argument shrinks). This is one option among multiple stability mechanisms (a $\lambda$-warmup schedule, gradient clipping, or smaller $\lambda$).

\paragraph{Application to which heads.}
We apply $\mathcal{L}_{\text{jac}}$ to position and scale heads only; color and opacity remain view-dependent. The residual head's input vector does not include the harmonic view encoding $\gamma(\mathbf{d})$ (App.~\ref{sec:method-app-arch}), so it is structurally independent of $\gamma(\mathbf{d})$ and is not penalized by $\mathcal{L}_{\text{jac}}$. Rotation is excluded because the local quaternion is constrained to the unit sphere; a Frobenius penalty on the raw 4-d quaternion components does not equal angular-Jacobian energy under the non-trivial Riemannian metric, making a Hutchinson-projection penalty ill-posed for $\mathbf{q}^{\text{local}}_g$.
\subsection{Per-subject optimization details}
\label{sec:method-app-opt}

\paragraph{Log-scale hard clamp and screen-space soft penalty.}
The hard clamp operates in parameter space: each iteration the log-scale tensor $\mathbf{s}^{\log}$ is clamped to a fixed absolute range $[s^{\log}_{\min}, s^{\log}_{\max}]$, then the per-Gaussian deviation from the log-scale mean is clamped so that $\max_a(\mathbf{s}_{g,a}^{\log}) - \min_a(\mathbf{s}_{g,a}^{\log})$ stays within a fixed log-spread bound across the three axes; this provides a hard upper bound on world-space anisotropy. The soft penalty $\mathcal{L}_{\text{aspect}}$ operates in screen space. The per-Gaussian world-space covariance $\Sigma^{\text{world}}_g = R^{\text{world}}_g \,\mathrm{diag}\big(({\mathbf{s}^{\text{world}}_g})^2\big)\, (R^{\text{world}}_g)^\top$ is projected to the camera by $\Sigma^{\text{2D}}_g = J_g\, W\, \Sigma^{\text{world}}_g\, W^\top J_g^\top$, where $W$ is the world-to-camera rotation and $J_g$ is the projection Jacobian at the Gaussian's center, following the standard 3DGS projection~\citep{kerbl20233d}. The closed-form $2\!\times\!2$ eigenvalues $\lambda_{\max}, \lambda_{\min}$ of $\Sigma^{\text{2D}}_g$ define the projected log-aspect $\tfrac{1}{2}(\log\lambda_{\max} - \log\lambda_{\min})$, which $\mathcal{L}_{\text{aspect}}$ penalizes above $\tau$. The two controls operate on different quantities (the hard clamp on parameter-space log-scale, the soft penalty on screen-space projected aspect) and are therefore not redundant: the hard clamp guarantees an absolute world-space bound, while the soft penalty supplies a smooth perceptually-aligned gradient toward isotropic projected footprints.
\paragraph{Dual-side background compositing.} The rendered image $\hat{\mathbf{I}}$ exits the differentiable 32-channel splatter together with a per-pixel splatting alpha
\begin{equation}
\hat{\alpha}(u) \;=\; 1 - \prod_{i \in \mathcal{N}(u)} \big(1 - \alpha_i\, G_i(u)\big),
\label{eq:splat-alpha}
\end{equation}
where $\alpha_i$ is the per-Gaussian opacity, $G_i(u)$ is the projected 2D Gaussian footprint at pixel $u$, and $\mathcal{N}(u)$ is the set of Gaussians overlapping $u$. This is the standard 3DGS opacity accumulation~\citep{kerbl20233d}, equivalent to one minus the unoccluded transmittance through $\mathcal{N}(u)$; the product is order-independent (front-to-back ordering matters for color compositing but not for the alpha-only product). The matte $\mathbf{m}$ is produced by the StyleMatte branch of the EMICA tracker (\texttt{xg-chu/GAGAvatar\_track}; see App.~\ref{sec:method-app-arch}) as in the feed-forward stage. We use \emph{premultiplied alpha compositing} on both sides:
\begin{equation}
\hat{\mathbf{I}}_{\text{comp}} = \mathbf{m} \odot \hat{\mathbf{I}} + (1 - \mathbf{m}) \odot \mathbf{b}, \qquad \mathbf{I}_{\text{comp}} = \mathbf{m} \odot \mathbf{I} + (1 - \mathbf{m}) \odot \mathbf{b}, \qquad \mathbf{b} \sim \mathcal{U}([0, 1]^3),
\label{eq:compositing}
\end{equation}
with $\mathbf{b}$ resampled at every iteration. We trace where $\mathbf{b}$ enters the gradient.

\emph{(i) Pure-$L_1$ in pixels where $\hat\alpha = \mathbf{m} = 1$.} The $L_1$ pixel residual reduces to $\|\hat{\mathbf{I}}_{\text{comp}} - \mathbf{I}_{\text{comp}}\|_1 = \|\mathbf{m}\odot(\hat{\mathbf{I}} - \mathbf{I})\|_1$ on the matted target side. On the rendered side, the splatter's actual output is $\hat\alpha\hat{\mathbf{I}} + (1-\hat\alpha)\mathbf{b}$ rather than $\mathbf{m}\hat{\mathbf{I}} + (1-\mathbf{m})\mathbf{b}$; the random $\mathbf{b}$ cancels exactly only at pixels where $\hat\alpha = \mathbf{m} = 1$ simultaneously. Pixels where $\hat\alpha \neq \mathbf{m}$ are handled by mechanism (iii) below.

\emph{(ii) SSIM-mediated boundary band.} The SSIM term in $\mathcal{L}_{\text{ref}}$ is computed on a Gaussian window of size $w\!\times\!w$ ($w=11$ as in the standard SSIM implementation). Because the window crosses the silhouette, the local-window means $\mu_{\hat{\mathbf{I}}_{\text{comp}}}, \mu_{\mathbf{I}_{\text{comp}}}$ and variances depend on $\mathbf{b}$ in a band of width $\lfloor w/2\rfloor$ pixels around the silhouette, letting the randomization take effect at and near the boundary.
\emph{(iii) $\hat\alpha\!-\!\mathbf{m}$ disagreement at the boundary.} When the splatter's $\hat{\alpha}$ disagrees with the matte $\mathbf{m}$ (silhouette overspill or partial transparency), the realized rendering at pixel $u$ is $C(u) = \hat{\alpha}(u)\hat{\mathbf{I}}(u) + (1{-}\hat{\alpha}(u))\mathbf{b}$. Treating $\mathbf{m}\mathbf{I} + (1-\mathbf{m})\mathbf{b}$ as the supervision target $y(u)$, the $L_1$ pixel loss $\|y - C\|_1$ has gradient $\partial \|y-C\|_1 / \partial \mathbf{b} = -(1-\hat{\alpha})\,\mathrm{sign}(y - C)$, which is non-zero on partially transparent pixels and switches sign with $\mathbf{b}$; the same $\mathbf{b}$ on both sides yields a \emph{symmetric} (rather than asymmetric) gradient that pulls $\hat{\alpha}\to\mathbf{m}$ without an explicit silhouette loss. The dual-side trick of using the same random $\mathbf{b}$ on both sides, a long-standing differentiable-rasterization technique~\citep{mildenhall2020nerf, mueller2022instant}, is here applied with $\hat{\alpha}\!-\!\mathbf{m}$ symmetrization in the per-subject Gaussian-refinement context.

\paragraph{Reference-frame selection.}
We initialize per-subject refinement from $f_\theta$'s output at a reference frame $r$ chosen by minimum jaw-pose magnitude, $r = \arg\min_t \|\boldsymbol{\theta}^{\text{jaw}}_t\|_1$, with ties broken by earliest frame index. If the minimum jaw-pose magnitude across all $T$ frames is below a noise threshold $\tau_{\text{jaw}}\!=\!10^{-3}$ (silent monologue, occluded jaw, or severe-pose-only video), we fall back to a uniformly random frame from the bottom-decile pool $\{t : \|\boldsymbol{\theta}^{\text{jaw}}_t\|_1 \leq \mathrm{quantile}_{10\%}\}$. This is a jaw-only heuristic and does not select a fully canonical pose.

\paragraph{Per-parameter learning rates and schedule.}
Per-parameter learning rates are tuned individually for each Gaussian attribute family (position, rotation, color, opacity, and scale, the last unfrozen at iteration $500$), with a single global cosine decay over the full ten thousand iterations. We omit LPIPS at this iteration budget and restrict it to the feed-forward stage: at $10$K steps with the per-parameter learning rates above, the LPIPS gradient produces VGG-edge-filter speckle artifacts on the frozen Gaussian set before convergence; the $L_1{+}\text{SSIM}$ combination recovers high-frequency content directly via Gaussian color and position.

\section{Additional experimental results}
\label{sec:exp-app-extras}

\subsection{Per-subject monocular qualitative comparison}
\label{sec:exp-app-persubj-qual}

\begin{figure}[t]
\centering
\includegraphics[width=0.96\textwidth]{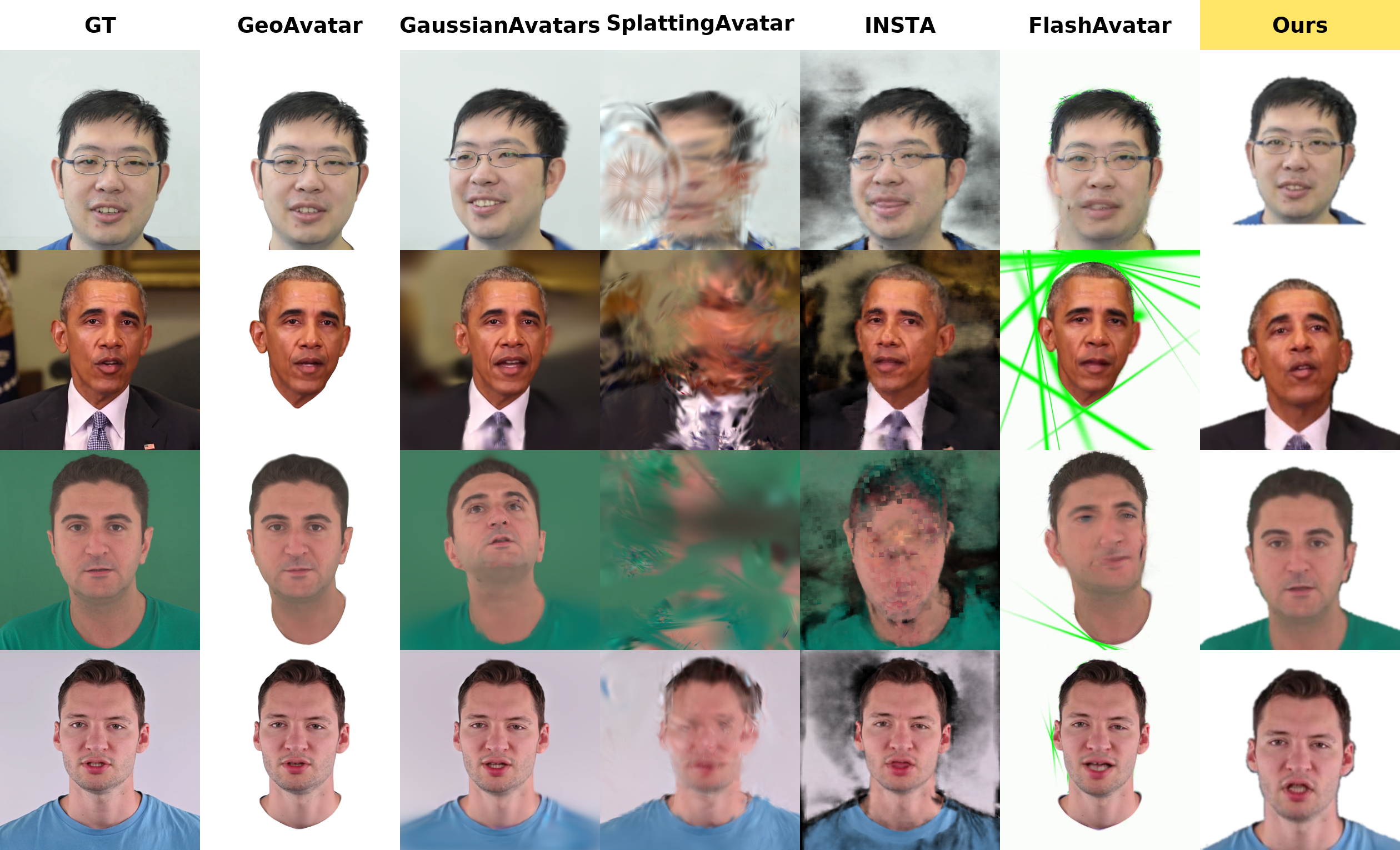}
\caption{\textbf{Per-subject monocular qualitative comparison
(visualization sample of $4$ subjects; full numerics in
Tab.~\ref{tab:main-persubj}).} Each row shows one held-out target
frame; columns are GT, GeoAvatar~\citep{moon2025geoavatar},
GaussianAvatars~\citep{qian2024gaussianavatars},
SplattingAvatar~\citep{shao2024splattingavatar},
INSTA~\citep{zielonka2023insta},
FlashAvatar~\citep{xiang2024flashavatar},
and \textbf{Ours+S3} (highlighted).}
\label{fig:qual-persubj}
\end{figure}

Fig.~\ref{fig:qual-persubj} shows qualitative renders accompanying the
per-subject monocular numerics of Tab.~\ref{tab:main-persubj} on a
$4$-subject visualization sample.

Each row places a held-out target frame next to the six candidate
renders (five per-subject baselines and Ours+S3). Ours+S3 follows
the target head pose closely with no visible blur or dropped
regions, and subject-specific high-frequency content (forehead
wrinkles, malar contour, beard or mustache texture, hair-edge
silhouette, eye specular highlights) remains sharp at the rendered
scale.

\subsection{Cross-identity feed-forward qualitative comparison}
\label{sec:exp-app-cross-id}

\begin{figure*}[t!]
\centering
\includegraphics[width=0.96\textwidth]{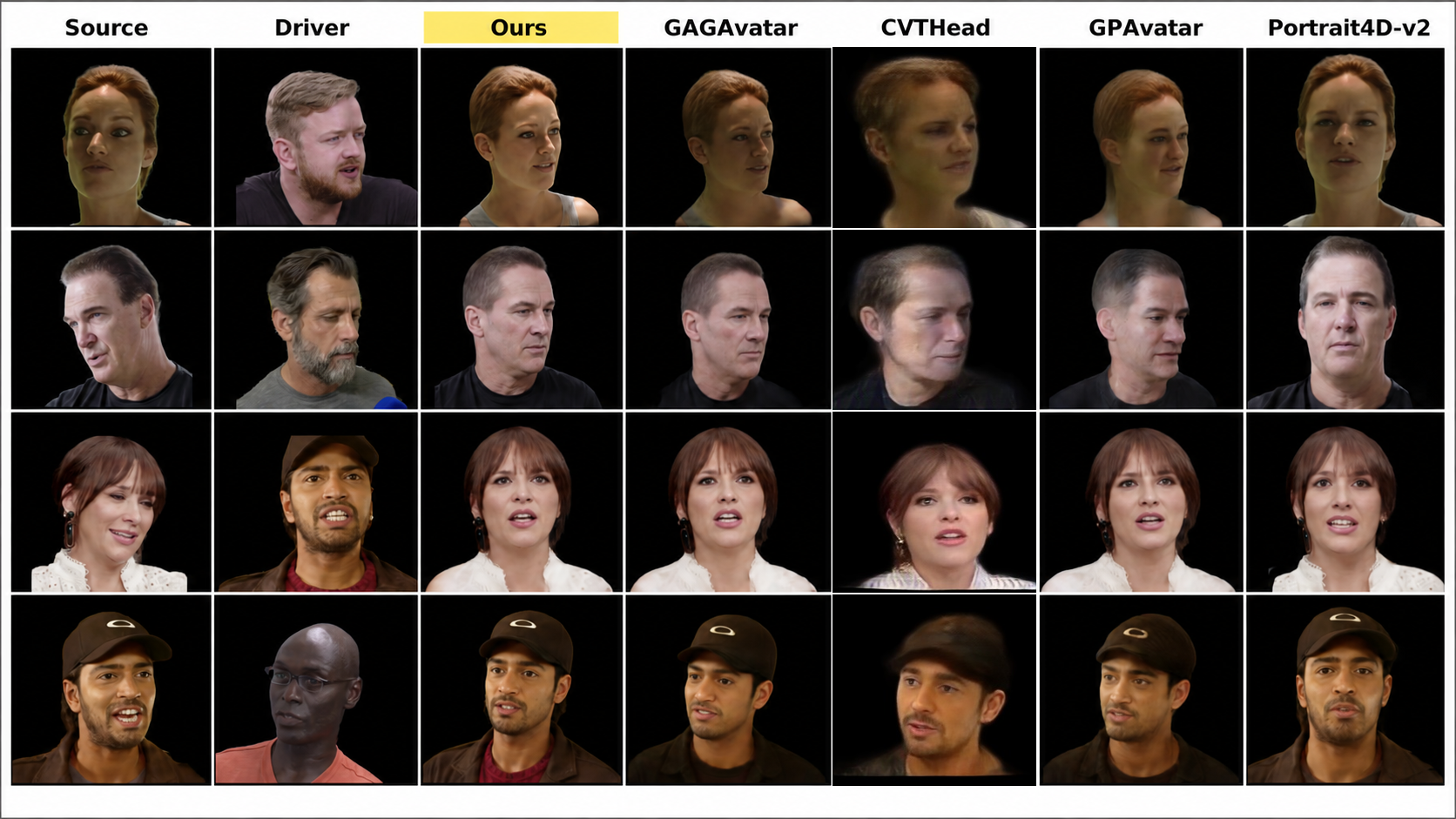}\\[2pt]
{\footnotesize (a) VFHQ test split (4 stratified-random cross-identity pairs)}\\[6pt]
\includegraphics[width=0.96\textwidth]{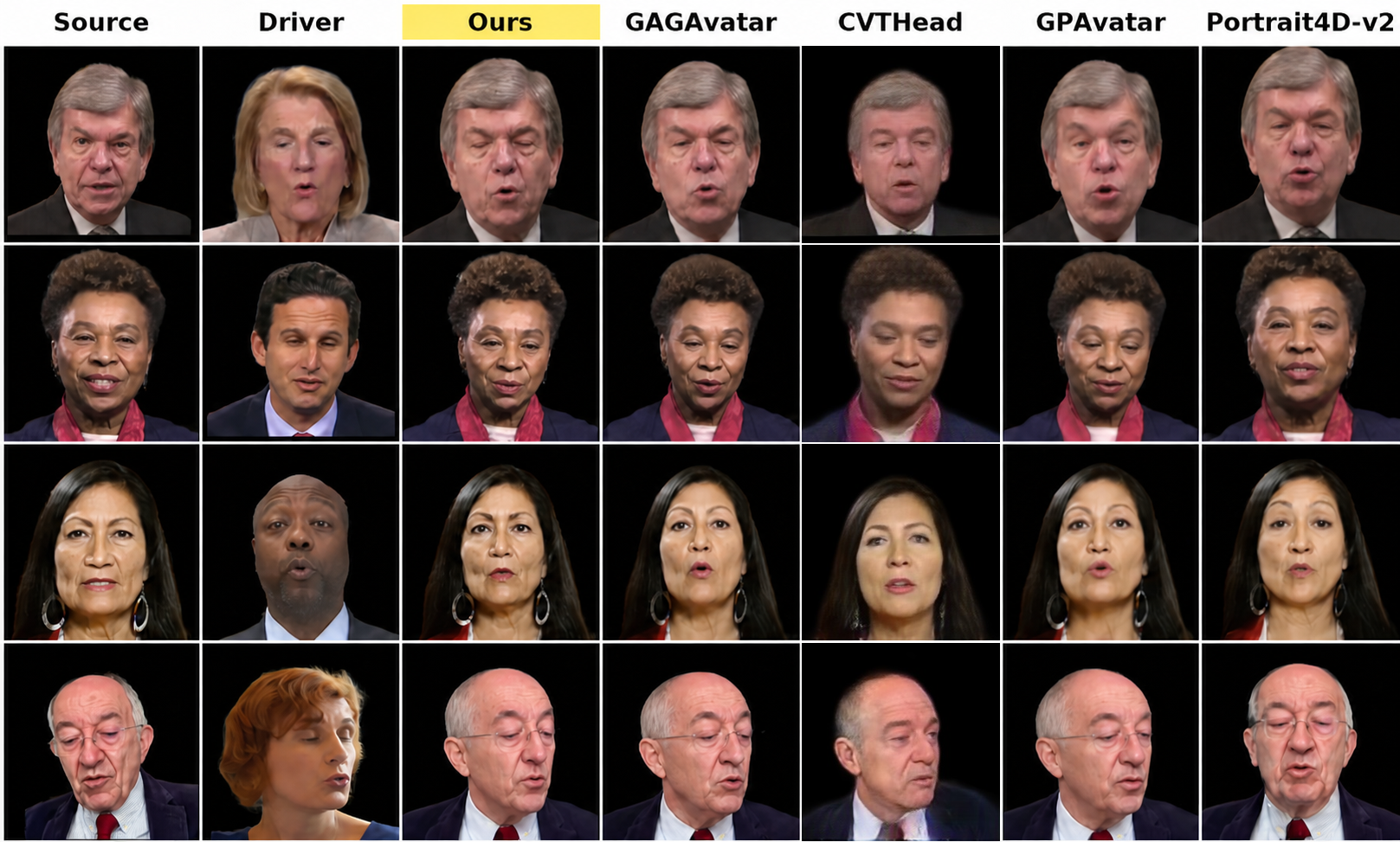}\\[2pt]
{\footnotesize (b) HDTF test split (4 stratified-random cross-identity pairs, GPAvatar split)}
\caption{\textbf{Cross-identity feed-forward qualitative comparison
(Tabs.~\ref{tab:main-ff-vfhq}--\ref{tab:main-ff-hdtf}, Cross block,
columns~9--11).} Each row pairs an identity-A source (col 1) with an
identity-B driver (col 2); the remaining columns are reenactments of
identity A under driver B's expression and pose. \textbf{(a)} VFHQ test
split, \textbf{(b)} HDTF test split. Each panel: 4 stratified-random
disjoint-identity pairs $\times$ 7 columns (source-A, driver-B,
\textbf{Ours-FF} (highlighted), GAGAvatar~\citep{chu2024generalizable},
CVTHead~\citep{ma2024cvthead}, GPAvatar~\citep{chu2024gpavatar}, and
Portrait4D-v2~\citep{deng2024portrait4dv2}). Real3D-Portrait and
Portrait4D dropped on the same grounds as
Fig.~\ref{fig:qual-cross-domain}.}
\label{fig:qual-cross-id}
\end{figure*}

The cross-block metrics (CSIM/AED/APD) of
Tabs.~\ref{tab:main-ff-vfhq}--\ref{tab:main-ff-hdtf} characterize the
cross-identity lane numerically; Fig.~\ref{fig:qual-cross-id}
visualizes whether high CSIM corresponds to recognizable identity
transfer or to a low-frequency average.

The cross-identity lane (Fig.~\ref{fig:qual-cross-id}) places
source-driver pairs whose head yaw and expression states differ
substantially. The second VFHQ row pairs a profile-view source with
a near-frontal driver under a strong open-mouth expression, and
analogous yaw-and-expression discrepancies recur in the remaining rows.
Ours-FF retargets the source identity to the driver pose with the
geometric face structure (eye contour, nose-bridge orientation, jaw
outline) remaining consistent with the source identity even under
these extreme yaw and expression conditions.

\subsection{Backbone capacity and source-image count}
\label{sec:exp-app-k-scaling}

The top block of Tab.~\ref{tab:k-scaling} compares the three publicly
released DINOv3 variants; the encoder is
frozen, so backbone capacity affects feature quality and forward-pass
cost but not the trainable-parameter count of the pipeline. The bottom
block evaluates the source-image count $K\!\in\!\{1,2,3,4\}$ at
inference time on the single feed-forward checkpoint.

\begin{table}[t]
\centering
\caption{\textbf{Model-parameter ablation on CelebV-HQ.} Same self-reenactment
column structure as Tab.~\ref{tab:ablation}. Top block: backbone-capacity
comparison across the three publicly released DINOv3
variants (encoder frozen; bold row marks the default).
Bottom block: inference-time source-image count $K$ on the single
feed-forward checkpoint. Row~$K{=}4$ is the largest training-time
condition.}
\label{tab:k-scaling}
\setlength{\tabcolsep}{4pt}
\resizebox{0.8\linewidth}{!}{%
\begin{tabular}{l ccccccc}
\toprule
Variant
& PSNR\,$\uparrow$ & SSIM\,$\uparrow$ & LPIPS\,$\downarrow$
& CSIM\,$\uparrow$ & AED\,$\downarrow$ & APD\,$\downarrow$ & AKD\,$\downarrow$ \\
\midrule
DINOv3-S                                             & 24.48 & 0.863 & 0.083 & 0.917 & 0.095 & 0.064 & 2.296 \\
\textbf{DINOv3-B}                                    & 24.94 & 0.881 & 0.070 & 0.935 & 0.084 & 0.054 & 2.032 \\
DINOv3-L                                             & \textbf{25.02} & \textbf{0.884} & \textbf{0.067} & \textbf{0.941} & 0.085 & \textbf{0.053} & \textbf{1.943} \\
\midrule
$K{=}1$                                              & 23.01 & 0.806 & 0.126 & 0.828 & \textbf{0.079} & 0.101 & 3.098 \\
$K{=}2$                                              & 24.21 & 0.856 & 0.090 & 0.908 & 0.092 & 0.071 & 2.396 \\
$K{=}3$                                              & 24.75 & 0.869 & 0.079 & 0.914 & 0.082 & 0.058 & 2.118 \\
$K{=}4$                                              & 24.94 & 0.881 & \textbf{0.070} & 0.935 & 0.084 & 0.054 & 2.032 \\
\bottomrule
\end{tabular}%
}
\end{table}

The backbone comparison traces the expected monotone improvement from
DINOv3-S to DINOv3-L, with DINOv3-B on the knee of the
quality--parameter curve. The $K$-source comparison shows the largest
jump from $K{=}1$ to $K{=}2$ and flattens thereafter.

\subsection{Training hyperparameter ablation}
\label{sec:exp-app-training-hp}

Tab.~\ref{tab:training-hp} probes the sensitivity of the
feed-forward checkpoint to four training-time hyperparameters: the
Phase~1 learning-rate schedule, the Phase~1 base learning rate, the
Phase~2 encoder layer-wise LR decay $\gamma_{\text{LR}}$, and the
Jacobian penalty weight $\lambda_{\text{jac}}$. Each block varies
one hyperparameter with the remaining hyperparameters held at the
default; row~(a) reproduces Tab.~\ref{tab:ablation} row~(C), and
the other Default rows~(e),(h),(k) are independent reruns of the
same configuration.

\begin{table}[t]
\centering
\caption{\textbf{Phase~1 and Phase~2 training hyperparameter ablation
on CelebV-HQ.} Same self-reenactment column structure as
Tab.~\ref{tab:ablation}. Each block varies one training-time
hyperparameter with the others fixed at the default. Row~(a)
reproduces Tab.~\ref{tab:ablation} row~(C); the other Default
rows~(e),(h),(k) are independent reruns of the same configuration
and differ by small evaluation-order noise.}
\label{tab:training-hp}
\setlength{\tabcolsep}{4pt}
\resizebox{\columnwidth}{!}{%
\begin{tabular}{c l ccccccc}
\toprule
& Variant
& PSNR\,$\uparrow$ & SSIM\,$\uparrow$ & LPIPS\,$\downarrow$
& CSIM\,$\uparrow$ & AED\,$\downarrow$ & APD\,$\downarrow$ & AKD\,$\downarrow$ \\
\midrule
\multicolumn{9}{@{}l}{\emph{Phase~1 LR schedule}}\\
(a) & \textbf{LinearLR $1.0{\to}0.1$, $160$K+$40$K (Default)}        & \textbf{24.94} & \textbf{0.881} & \textbf{0.070} & \textbf{0.935} & 0.084          & \textbf{0.054} & \textbf{2.032} \\
(b) & cosine decay over $200$K iters                                 & 24.81          & 0.879          & 0.072          & 0.929          & \textbf{0.082} & 0.056          & 2.118          \\
(c) & constant LR ($2.5{\times}10^{-4}$)                             & 24.51          & 0.871          & 0.077          & 0.922          & 0.091          & 0.061          & 2.207          \\
\midrule
\multicolumn{9}{@{}l}{\emph{Phase~1 base LR}}\\
(d) & $1{\times}10^{-4}$ (under)                                     & 24.71          & 0.875          & 0.074          & 0.928          & \textbf{0.083} & 0.058          & 2.122          \\
(e) & \textbf{$2.5{\times}10^{-4}$ (Default)}                        & \textbf{24.92} & \textbf{0.880} & \textbf{0.071} & \textbf{0.934} & 0.085          & \textbf{0.055} & \textbf{2.041} \\
(f) & $5{\times}10^{-4}$ (over)                                      & 24.42          & 0.866          & 0.082          & 0.916          & 0.095          & 0.064          & 2.273          \\
\midrule
\multicolumn{9}{@{}l}{\emph{Phase~2 encoder layer-wise LR decay $\gamma_{\text{LR}}$}}\\
(g) & $\gamma_{\text{LR}}{=}0.5$ (more frozen)                       & 24.84          & 0.878          & 0.072          & 0.926          & 0.087          & 0.057          & \textbf{2.011} \\
(h) & \textbf{$\gamma_{\text{LR}}{=}0.65$ (Default)}                 & \textbf{24.96} & \textbf{0.882} & \textbf{0.069} & \textbf{0.937} & \textbf{0.083} & \textbf{0.054} & 2.024          \\
(i) & $\gamma_{\text{LR}}{=}0.8$ (less decay)                        & 24.80          & 0.878          & 0.071          & 0.928          & 0.090          & 0.057          & 2.058          \\
\midrule
\multicolumn{9}{@{}l}{\emph{Jacobian penalty weight $\lambda_{\text{jac}}$}}\\
(j) & $\lambda_{\text{jac}}{=}0$ (no penalty)                        & 24.30          & 0.876          & \textbf{0.067} & 0.929          & 0.092          & 0.063          & 2.301          \\
(k) & \textbf{$\lambda_{\text{jac}}{=}10^{-1}$ (Default)}            & \textbf{24.93} & \textbf{0.880} & 0.070          & \textbf{0.933} & \textbf{0.084} & \textbf{0.055} & 2.046          \\
(l) & $\lambda_{\text{jac}}{=}1$ (over-regularized)                  & 24.46          & 0.867          & 0.082          & 0.917          & 0.097          & 0.066          & \textbf{2.005} \\
\bottomrule
\end{tabular}%
}
\end{table}

The Phase~1 LR schedule rows~(a)--(c) show the LinearLR default
outperforming both cosine~(b) and constant~(c) by
${\sim}0.1$--$0.5$\,dB PSNR, with constant LR the worst. The base-LR
rows~(d)--(f) form a single peak around $2.5{\times}10^{-4}$ where
halving or doubling each lose ${\sim}0.2$--$0.5$\,dB. The Phase~2
layer-wise decay $\gamma_{\text{LR}}$~(g)--(i) is the least
sensitive setting on PSNR: deviations either way regress by under
$0.2$\,dB, while the L2-SP anchor and cross-time mix
(Tab.~\ref{tab:phase2-anchor-mix}) carry the Phase~2 cross-domain
protection. The $\lambda_{\text{jac}}$ rows~(j)--(l) form a sharp
peak at $\lambda_{\text{jac}}{=}10^{-1}$: setting it to $0$~(j) or
$1$~(l) each lose ${\sim}0.4$--$0.7$\,dB.

\subsection{L2-SP anchor and cross-time mix ablation}
\label{sec:exp-app-phase2-anchor-mix}

Tab.~\ref{tab:phase2-anchor-mix} probes the two Phase~2 controls introduced
in \S\ref{sec:method-feedforward}: the L2-SP anchor weight
$\lambda_{\text{sp}}$ and the NeRSemble cross-time mix share.
Rows~(a)--(d) form a $2{\times}2$ factorial over
$\lambda_{\text{sp}}\!\in\!\{0,10^{-3}\}$ and cross-time mix share
$\!\in\!\{0\%, 25\%\}$, isolating each control's individual
contribution; rows~(e)--(f) vary
$\lambda_{\text{sp}}\!\in\!\{10^{-2},10^{-4}\}$ at the chosen mix
share. Rows~(a) and (d) duplicate Tab.~\ref{tab:ablation} rows~(B) and
(C) for self-containment.

\begin{table}[t]
\centering
\caption{\textbf{L2-SP anchor and cross-time mix ablation on CelebV-HQ.} Same
self-reenactment column structure as Tab.~\ref{tab:ablation}. Top block
(rows~a--d): $2{\times}2$ factorial over the L2-SP anchor weight
$\lambda_{\text{sp}}$ and the NeRSemble cross-time mix share. Bottom
block (rows~e--f): $\lambda_{\text{sp}}$ sensitivity at the chosen mix
share. Row~(d) is the default Phase~2 configuration used elsewhere in
the paper. Rows~(a) and (d) duplicate Tab.~\ref{tab:ablation} rows~(B)
and (C).}
\label{tab:phase2-anchor-mix}
\setlength{\tabcolsep}{4pt}
\resizebox{\columnwidth}{!}{%
\begin{tabular}{c l ccccccc}
\toprule
& Variant
& PSNR\,$\uparrow$ & SSIM\,$\uparrow$ & LPIPS\,$\downarrow$
& CSIM\,$\uparrow$ & AED\,$\downarrow$ & APD\,$\downarrow$ & AKD\,$\downarrow$ \\
\midrule
(a) & $\lambda_{\text{sp}}{=}0$, cross-time $0\%$                                  & 22.73 & 0.799 & 0.135 & 0.814 & 0.117 & 0.112 & 3.244 \\
(b) & $\lambda_{\text{sp}}{=}10^{-3}$, cross-time $0\%$                            & 24.11 & 0.854 & 0.097 & \textbf{0.945} & 0.100 & 0.079 & 2.459 \\
(c) & $\lambda_{\text{sp}}{=}0$, cross-time $25\%$                                 & 23.81 & 0.840 & 0.104 & 0.868 & 0.107 & 0.082 & 2.573 \\
(d) & \textbf{$\lambda_{\text{sp}}{=}10^{-3}$, cross-time $25\%$ (Default)}        & \textbf{24.94} & \textbf{0.881} & \textbf{0.070} & 0.935 & 0.084 & \textbf{0.054} & \textbf{2.032} \\
\midrule
(e) & $\lambda_{\text{sp}}{=}10^{-2}$, cross-time $25\%$                           & 24.37 & 0.862 & 0.089 & 0.909 & \textbf{0.077} & 0.069 & 2.302 \\
(f) & $\lambda_{\text{sp}}{=}10^{-4}$, cross-time $25\%$                           & 24.52 & 0.869 & 0.077 & 0.916 & 0.090 & 0.063 & 2.283 \\
\bottomrule
\end{tabular}%
}
\end{table}

The $2{\times}2$ factorial shows that the anchor alone~(b) and
cross-time mix alone~(c) each provide partial cross-domain protection,
while the joint configuration~(d) recovers the full effect, indicating
the two mechanisms are complementary rather than redundant. The
$\lambda_{\text{sp}}$ rows~(e)--(f) show the expected
sensitivity: under-anchoring~(f) approaches cross-time-mix-only~(c),
while over-anchoring~(e) reduces Phase~2's ability to absorb
multi-view signal. We adopt $\lambda_{\text{sp}}{=}10^{-3}$ with
$25\%$ cross-time mix (row~d) as the default.

\subsection{Anti-spike component ablation}
\label{sec:exp-app-antispike}

Tab.~\ref{tab:antispike} probes the three components of the
per-subject refinement anti-spike regularization described in
\S\ref{sec:method-optimization}: the
scale-freeze warmup over iterations $0$--$500$, the per-iteration
log-scale clamp applied after iteration $500$, and the screen-space
anti-anisotropy penalty $\mathcal{L}_{\text{aspect}}$ active from
iteration $500$. Row~(a) is the full configuration (matching
Tab.~\ref{tab:ablation} row~(D)); rows~(b)--(d) each remove one
component while holding the other two fixed.

\begin{table}[t]
\centering
\caption{\textbf{Per-subject anti-spike component ablation on CelebV-HQ.}
Same self-reenactment column structure as Tab.~\ref{tab:ablation}.
Row~(a) is the full anti-spike regularization (matches Tab.~\ref{tab:ablation}
row~(D)); rows~(b)--(d) each remove one component, holding the other
two fixed.}
\label{tab:antispike}
\setlength{\tabcolsep}{4pt}
\resizebox{\columnwidth}{!}{%
\begin{tabular}{c l ccccccc}
\toprule
& Variant
& PSNR\,$\uparrow$ & SSIM\,$\uparrow$ & LPIPS\,$\downarrow$
& CSIM\,$\uparrow$ & AED\,$\downarrow$ & APD\,$\downarrow$ & AKD\,$\downarrow$ \\
\midrule
(a) & \textbf{Full anti-spike (scale-freeze + log-clamp + $\mathcal{L}_{\text{aspect}}$)} & \textbf{29.95} & \textbf{0.947} & 0.045 & \textbf{0.959} & 0.057 & 0.039 & \textbf{1.365} \\
\midrule
(b) & (a) w/o scale-freeze warmup                                              & 27.88 & 0.941 & 0.070 & 0.945 & \textbf{0.051} & 0.043 & 1.623 \\
(c) & (a) w/o log-scale clamp                                                  & 28.39 & 0.938 & \textbf{0.043} & 0.947 & 0.063 & 0.038 & 1.644 \\
(d) & (a) w/o $\mathcal{L}_{\text{aspect}}$ ($\lambda_{\text{aspect}}{=}0$)    & 28.92 & 0.930 & 0.051 & 0.954 & 0.062 & \textbf{0.035} & 1.579 \\
\bottomrule
\end{tabular}%
}
\end{table}

Removing the scale-freeze warmup~(b) restores early-iteration
anisotropy spikes that $\mathcal{L}_{\text{aspect}}$ alone (active
after iter $500$) cannot retroactively arrest. Removing the
log-clamp~(c) lets accumulated log-scale drift past the budget between
iterations, which $\mathcal{L}_{\text{aspect}}$ as a soft penalty
cannot fully reabsorb. Removing
$\mathcal{L}_{\text{aspect}}$~(d) lets gaussians regrow into thin
screen-space streaks once the projection direction shifts away from
training views, against which the warmup-then-clamp schedule provides
no view-dependent protection. These failure modes do not overlap: the
full regularization keeps $\max\!\mathrm{AR}\,{\approx}\,8$ at iter $10$K,
removing $\mathcal{L}_{\text{aspect}}$ drifts to ${\approx}\,20$, and
removing the log-clamp causes catastrophic elongation
($\max\!\mathrm{AR}\,{>}\,10^{6}$, runtime status=1). The refinement
photometric loss largely compensates at the rendered output (PSNR
spread under $2.1$~dB), so $\max\!\mathrm{AR}$ rather than rendered
appearance is the more informative signal: each of the three
components addresses a distinct failure mode.

\section{Limitations and Future Work}
\label{sec:limitations}

Our method inherits two limitations from its FLAME-mesh foundation:
large-scale accessories such as glasses or hats are not explicitly
modeled and may be folded into the head's surface texture, and the
upstream FLAME tracker assumes a successful frontal face detection on
at least one source frame. The per-subject refinement loop also remains
a ${\sim}2$-minute step, so true real-time avatar creation is restricted
to the feed-forward stage. Two natural extensions follow:
(i)~coupling with audio- and motion-driven control signals to form a
complete digital-human stack beyond the still-portrait input regime,
and
(ii)~extending the FLAME-mesh-bound Gaussian representation with body
and hand sub-meshes for full-body 4D digital humans.